\documentclass[lettersize,journal]{IEEEtran}
\usepackage{amsmath,amsfonts}
\usepackage{algorithmic}
\usepackage{array}
\usepackage[caption=false,font=normalsize,labelfont=sf,textfont=sf]{subfig}
\usepackage{textcomp}
\usepackage{stfloats}
\usepackage{url}
\usepackage{verbatim}
\usepackage{graphicx}
\usepackage{footnote}
\usepackage{longtable}
\def\BibTeX{{\rm B\kern-.05em{\sc i\kern-.025em b}\kern-.08em
    T\kern-.1667em\lower.7ex\hbox{E}\kern-.125emX}}
\usepackage{balance}
\usepackage{cite}
\usepackage{tabularx}
\usepackage{multirow}
\usepackage{makecell}
\usepackage{pifont}
\usepackage{titlesec}
\titlespacing{\section}{0pt}{2.5ex plus .2ex minus .2ex}{1.0ex plus .2ex}
\begin{document}
\newcommand{\e}[1]{\textbf{#1}}
\title{FD-RIO: Fast Dense Radar Inertial Odometry}
\author{Nader J. Abu-Alrub, Nathir A. Rawashdeh,~\IEEEmembership{Senior Member,~IEEE}
\thanks{Nader J. Abu-Alrub is with the Department of Applied Computing, Michigan Technological University.}
\thanks{Nathir A. Rawashdeh is with the Department of Applied Computing, Michigan Technological University.}
}

\maketitle

\begin{abstract}
Radar-based odometry is a popular solution for ego-motion estimation in conditions where other exteroceptive sensors may degrade, whether due to poor lighting or challenging weather conditions; however, scanning radars have the downside of relatively lower sampling rate and spatial resolution. In this work, we present FD-RIO, a method to alleviate this problem by fusing noisy, drift-prone, but high-frequency IMU data with dense radar scans. To the best of our knowledge, this is the first attempt to fuse dense scanning radar odometry with IMU using a Kalman filter. We evaluate our methods using two publicly available datasets and report accuracies using standard KITTI evaluation metrics, in addition to ablation tests and runtime analysis. Our phase correlation -based approach is compact, intuitive, and is designed to be a practical solution deployable on a realistic hardware setup of a mobile platform. Despite its simplicity, FD-RIO is on par with other state-of-the-art methods and outperforms in some test sequences.
\end{abstract}

\begin{IEEEkeywords}
Radar, IMU, Odometry, Sensor Fusion, Kalman Filter.
\end{IEEEkeywords}

\section{Introduction}\label{sec:introduction}
\IEEEPARstart{O}{dometry} 
can be implemented using a variety of sensors. It can be based on wheel encoders, Inertial Measurement Units (IMU), cameras (i.e., visual odometry), lidars, and radars. Radar-based odometry is of particular interest in the domain of robotics and autonomous vehicles; they are more resilient to various weather and lighting conditions, they are unaffected by slip and skid errors, the automotive type radars are relatively cheap, and they are being deployed in autonomous vehicles for a variety of applications such as adaptive cruise control and assisted parking. Following the taxonomy presented in \cite{NaderSurvey}, radar odometry methods can be classified as sparse (or indirect methods), dense (or direct methods), and hybrid methods. Generally, sparse methods use detections or distributions and rely on matching them between radar measurements. Dense methods process full scans to generate state estimates. Lastly, hybrid methods use a combination of both approaches.

The majority of radar odometry research is focused on sparse methods \cite{NaderSurvey} due to their presumed computational efficiency. A typical sparse radar odometry pipeline begins with preprocessing the data and extracting features. It then uses those features (or landmarks) to estimate a transformation that describes the pose change between two scans of interest. Additionally, there could be more intermediate steps used to enhance the overall performance, such as keyframe selection, factor graph optimization, outlier rejection (RANSAC or maximum clique), or motion distortion compensation. 

Dense odometry methods can be based on cross-correlation, deep learning, or phase correlation. Despite their relative simplicity, they are considerably underexplored compared to sparse methods. Dense methods are preferred in featureless environments (e.g., rural driving, mines, tunnels, etc.) where many feature-based methods perform poorly. They can also be leveraged to skip many of the computationally expensive intermediate stages necessary for sparse methods, such as motion compensation and outlier rejection. Additionally, dense methods eliminate all the downsides of relying on detection of landmarks, tracking, or matching. The reason why dense radar odometry methods are often overlooked could be attributed to the assumption that since dense radar odometry methods draw inspiration from dense visual odometry (commonly referred to as direct or appearance-based methods), they probably inherit the same weaknesses. Typical disadvantages of dense visual odometry methods include sensitivity to sudden changes in lighting and illumination conditions, dependency on good initialization in the case of optical flow -based methods, and computational cost. We demonstrate in this work that this assumption is not necessarily true; radars are unaffected by lighting and illumination, dense radar odometry methods can be implemented without the need to track the optical flow in their image-like scans, and lastly, dense radar scans can be processed in grayscale and in much smaller sizes compared to their visual counterparts.

Recent trends in radar odometry research have been leaning toward increased complexity, often disregarding the practical aspects of radar odometry; an algorithm that is supposed to run in real-time on usually limited computational resources. It is highly unlikely that a mobile platform will carry multiple high-end processors with multiple GPUs on board. It is also likely that other heavier algorithms are expected to run on the same hardware (e.g., object detection, tracking, mapping ... etc.). We advocate compact and elegant solutions and take a step forward in this direction, we propose a method that is simple and capable of running in real-time on a modest hardware setup, yet its performance is on par with the state-of-the-art methods that incorporate much more complexity. This is achieved through careful selection of classical and well-proven techniques in addition to stripping down intermediate components to their bare minimum. We argue that slight improvement in accuracy at the cost of encumbered hardware is hardly justifiable in the real world for mobile vehicles with constrained computational and energy resources.

In brief, the main contributions of this work can be summarized as follows:
\begin{itemize}
    \item We propose FD-RIO (Fast Dense Radar Inertial Odometry), a fast and efficient dense radar odometry algorithm with performance that is on par with other comparable methods. 
    \item To the best of our knowledge, FD-RIO is the only published attempt to fuse scanning radar and IMU data in a Kalman Filter framework.
    \item We evaluate our method on two publicly available scanning radar datasets. In addition to an ablation study and runtime analysis.
\end{itemize}

The remainder of this article is organized as follows, Section \ref{sec:related_work} summarizes the recent work on scanning radar odometry and radar-inertial odometry for mobile ground vehicles. Section \ref{sec:proposed_method} discusses the details of the proposed method and its implementation. Section \ref{sec:results} presents an evaluation of the proposed method along with experimental results and ablation tests.  Finally, section \ref{sec:conclusion} concludes this article and suggests some improvements and future work.

\section{Related Work}\label{sec:related_work}
The seminal work of Cen and Newman \cite{cen2018precise} is among the earliest publications on scanning radar odometry. Their proposed method detects and matches features extracted from 1D radial signals of a scanning radar, then they use Singular Value Decomposition (SVD) to solve for the transformation that aligns the scans. The same researchers later improved their feature detection and association in \cite{cen2019radar}. Burnett \textit{et al.} \cite{burnett2021mcransac} presented MC-RANSAC, short for Motion-Compensated RANdom SAmple Consensus. Their method relies on detecting and matching Oriented Fast and Rotated BREIF (ORB) descriptors between radar scans and using an optimizer to estimate the velocity while taking motion distortion into consideration. Burnett \textit{et al.} \cite{burnett2021radar} also proposed HERO, Hybrid-Estimate Radar Odometry, where they take a probabilistic approach for states estimate using Exactly Sparse Gaussian Variational Inference (ESGVI) and combine it with a Convolutional Neural Network (CNN) model for feature detection. Hong \textit{et al.} \cite{hong2022radarslam} presented RadarSLAM, its odometry pipeline relies on detecting blobs in radar scans and tracking them into the next scan. It then estimates the transformation using SVD. Additionally, they compensate for motion distortion and optimize the estimates using factor graph. Adolfsson \textit{et al.} \cite{adolfsson2023cfear} introduced CFEAR, an abbreviation for Conservative Filtering for Efficient and Accurate Radar odometry, which is a Point-to-Normal scan matching method that begins with filtering radar scans by keeping a preset number of strongest radar returns. The kept points are then used to form surfaces and surface normals. Their method relies on scan-to-keyframe matching and compensates for motion distortion. Zhang \textit{et al.} \cite{zhang2023sdr} developed an algorithm to denoise radar scans and filter out speckle noise and ghost objects. As for motion estimation, they use an improved Normal Distribution Transform (NDT) scan matching. Lim \textit{et al.} \cite{lim2023orora} presented ORORA, a method based on the work presented in \cite{burnett2021mcransac} with the addition of a new outlier rejection step, namely, Max Clique Inlier Selection (MCIS).

More related to the method we are proposing here are dense radar odometry methods. Lu \textit{et al.} \cite{lu2020milliego} projected scans of automotive radar into 2D images and fed them to a CNN model. Parallel to that, IMU measurements are processed by a Recurrent Neural Network (RNN), and the output of both pipelines are fused using another RNN model that generates estimates based on previous poses. Barnes \textit{et al.} \cite{barnes2019masking} proposed a method called masking by moving, where they also used a CNN model to generate masks. These masks are used to filter out noise and artifacts from radar scans. The filtered scans are then used for pose estimation based on cross-correlation. The rotation step is iterative and initially involved searching for the best rotation. This search process was later improved by Weston \textit{et al.} \cite{weston2022fastMbyM} where the rotation estimation step was replaced with phase correlation on polar coordinates of the magnitudes of Fourier Transform.

A branch of dense radar odometry groups methods that are based on phase correlation techniques, this category encompasses the work of Checchin \textit{et al.} \cite{checchin2010radar}, Park \textit{et al.} \cite{park2020pharao}, and Lubanco \textit{et al.} \cite{r3o}, in addition to our proposed method FD-RIO. Checchin \textit{et al.} \cite{checchin2010radar} proposed one of the earliest dense radar odometry methods that is based on the image registration technique introduced in \cite{fft_reddy} where phase correlation is used to estimate the translation and rotation between two consecutive radar scans on two stages. This early attempt, however, was designed for static environments and was not evaluated on a publicly available dataset. Park \textit{et al.} \cite{park2020pharao} developed a similar method that relies on phase correlation to estimate translation and rotation using separate stages; however, their method breaks the translation estimation step further down into two steps, one using downsampled scans and another using full resolution to refine the estimation. They improve the performance by implementing pose graph optimization and keyframe selection. Lubanco \textit{et al.} \cite{r3o} developed a closely related approach named $R^3O$ where they used the Radon transform for improved rotation estimates. Their method includes an outlier rejection, keyframe selection, and graph optimization steps in order to enhance the overall accuracy.

Using Kalman filter with automotive type radars for state estimation and fusion is not uncommon. Almalioglu \textit{et al.} \cite{almalioglu2020milli} developed \textit{Milli-RIO}, where they used NDT scan matching of automotive radar scans aided by integrated IMU data from an Unscented Kalman Filter (UKF); however, the UKF was not leveraged for fusion, and their implementation relies on an RNN model on top of the UKF and NDT matching pipelines to benefit from previous poses and improve the performance. de Araujo \textit{et al.} \cite{de2023novel} fused scans from an automotive radar along with IMU data using the Invariant Extended Kalman Filter (IEKF) to estimate poses, their algorithm depends on denoising the radar data using an autoencoder neural network. Holder \textit{et al.} \cite{holder2019real} proposed a method based on UKF to fuse preprocessed automotive radar data with measurements taken from the vehicle's Controller Area Network (CAN) bus, including the steering angle, gyroscope, and wheel encoders readings. Finally, Liang \textit{et al. } \cite{liang2020scalable} used the Invariant Extended Kalman Filter (IEKF) to fuse data from an automotive radar, Global Navigation Satellite System (GNSS), IMU, camera, and lidar. When compared to other Kalman and fusion-based implementations, our method stands out as the only attempt at using Kalman filter to fuse scanning radar data with IMU where others focused on fusing data from automotive radars. There are some fundamental differences between scanning radars and automotive radars that can affect the way they are fused with other sensors especially when using the Kalman filter approach. For automotive radars, the availability of radial velocities could be one the most important advantages; this velocity information is typically used to remove static objects from the scene in addition to state updates in the Kalman filter. Moreover, automotive radars have much faster sampling rates compared to scanning radars which could pose a challenge for scanning radars on higher driving speeds. Lastly, automotive radars are not prone to motion distortion; a problem stemming from the rotating mechanism of scanning radars. On the other hand, scanning radars have significantly richer scans that can be useful in a multitude of applications (e.g., state estimation, object detection, lane detection, etc.) whereas automotive radars have more dispersed returns.

Finally, FD-RIO is a model-based approach. It offers some potential advantages over training-based methods. For example, it does not require prolonged training during development. It is also generalizable; our tests show that FD-RIO performed well on two different datasets. Additionally, it is explainable and tunable; its various components and parameters can be easily modified or enhanced to expand it for different applications. 

\section{Proposed Method}\label{sec:proposed_method}
\subsection{Problem Statement}\label{subseq:problem}
The main goal of odometry is to estimate the states of a mobile robot along a certain trajectory. These states are usually the poses but can also include the velocities of the platform. We aim to estimate the homogeneous transformation $^{k-1}T_k \in \mathbb{R}^{4x4}$ that aligns two consecutive poses $P_{k-1}$ and $P_k$ where $k=\{0, 1, 2, ... n\}$ and $P_k$ is the pose at discrete instance $k$ and contains the position and orientation as in:
\begin{equation}
    \label{equ:pose_def}
    {P_k = }
    \begin{bmatrix}
        \textbf{R}^{3\times3}_k & \textbf{t}^{3\times1}_k  \\ \textbf{0}^{1\times3} & 1 
    \end{bmatrix}
\end{equation}
where $^{k-1}\textbf{R}_k \in SO(3)$ is the rotation matrix, and $^{k-1}\textbf{t}_k \in \mathbb{R}^{3x1}$ is the translation vector. After estimating all the transformations $^{k:n-1}T_{k+1:n}$, one can use concatenation as in equation (\ref{equ:poses_concatentaion}) to retrieve the full trajectory of the mobile platform using (\ref{equ:poses_set}).
\begin{equation}
    \label{equ:poses_concatentaion}
    ^0P_k =   ^0P_{k-1} \ ^{k-1}T_k
\end{equation}
\begin{equation}
    \label{equ:poses_set}
    ^0P_{k:n} = \{^0P_0, ^0P_1, ^0P_2, ... ^0P_n\}
\end{equation}

\subsection{Overview of FD-RIO}\label{subseq:overview}
\begin{figure}[!t]
    \centering
    \includegraphics[width=3.2in]{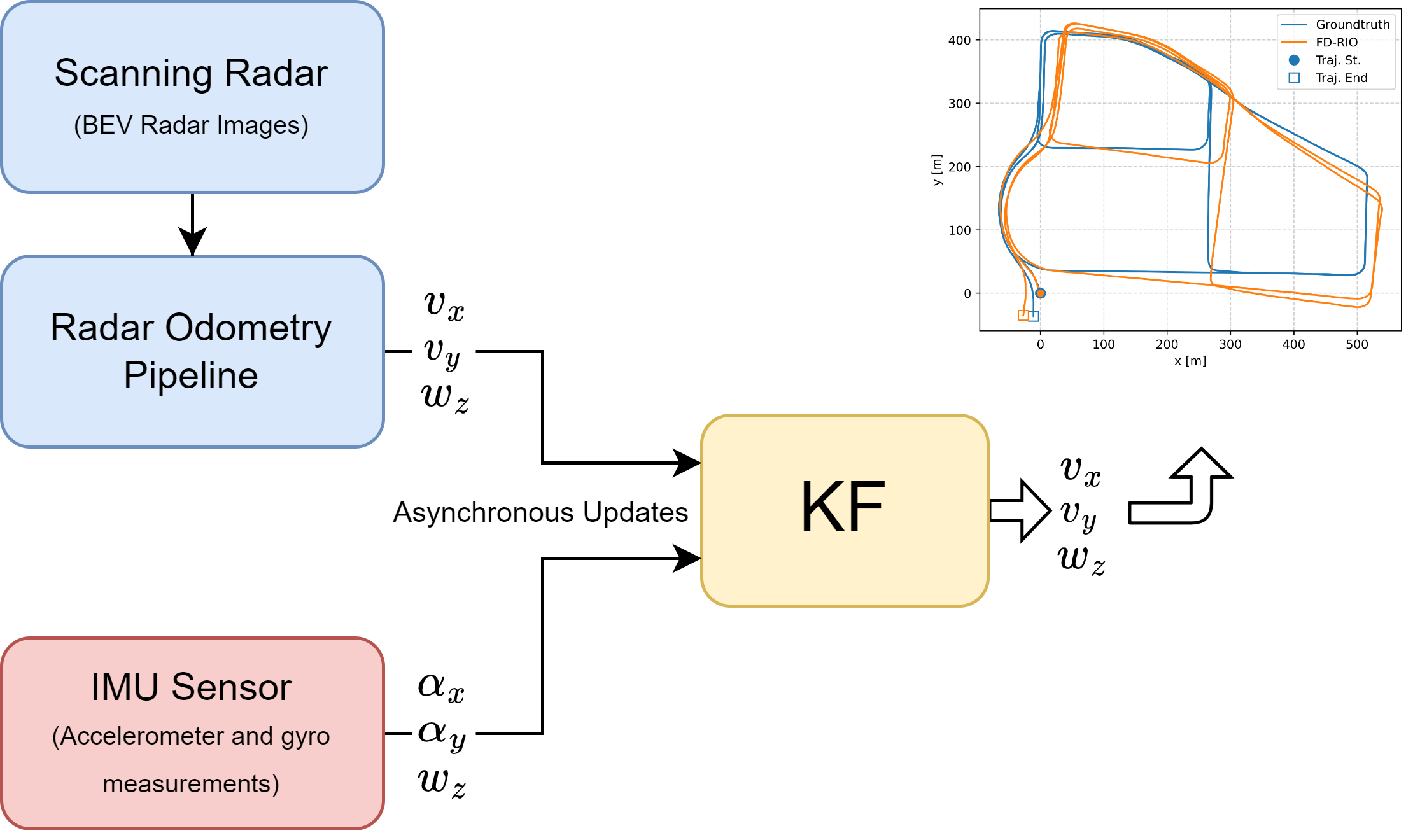}
    \caption{Overview of FD-RIO main components. Radar scans are processed by the radar odometry pipeline, its estimates are fed to a Kalman filter. IMU measurements are also sent to the filter where state estimates from both sensors are fused.}\label{fig:top_level}
\end{figure}

At the heart of FD-RIO is the Kalman filter which fuses data from two sources, the IMU sensor and the phase correlation -based radar odometry pipeline. This architecture is illustrated in Fig. \ref{fig:top_level}. Raw accelerometer and gyroscope readings are taken from the IMU and fed to the filter without any preprocessing. The readings of interest here are the linear acceleration in \textit{x} and \textit{y} dimensions and the angular velocity around \textit{z}. The other two major components of FD-RIO are the Kalman filter detailed in subsection \ref{subseq:kf} and the radar odometry pipeline in subsection \ref{subseq:pc_ro}. At the output end, the filter generates estimates of linear and angular velocities which in turn, are integrated and converted into poses.

\begin{figure*}[!t]
    \centering
    \includegraphics[width=7.0in]{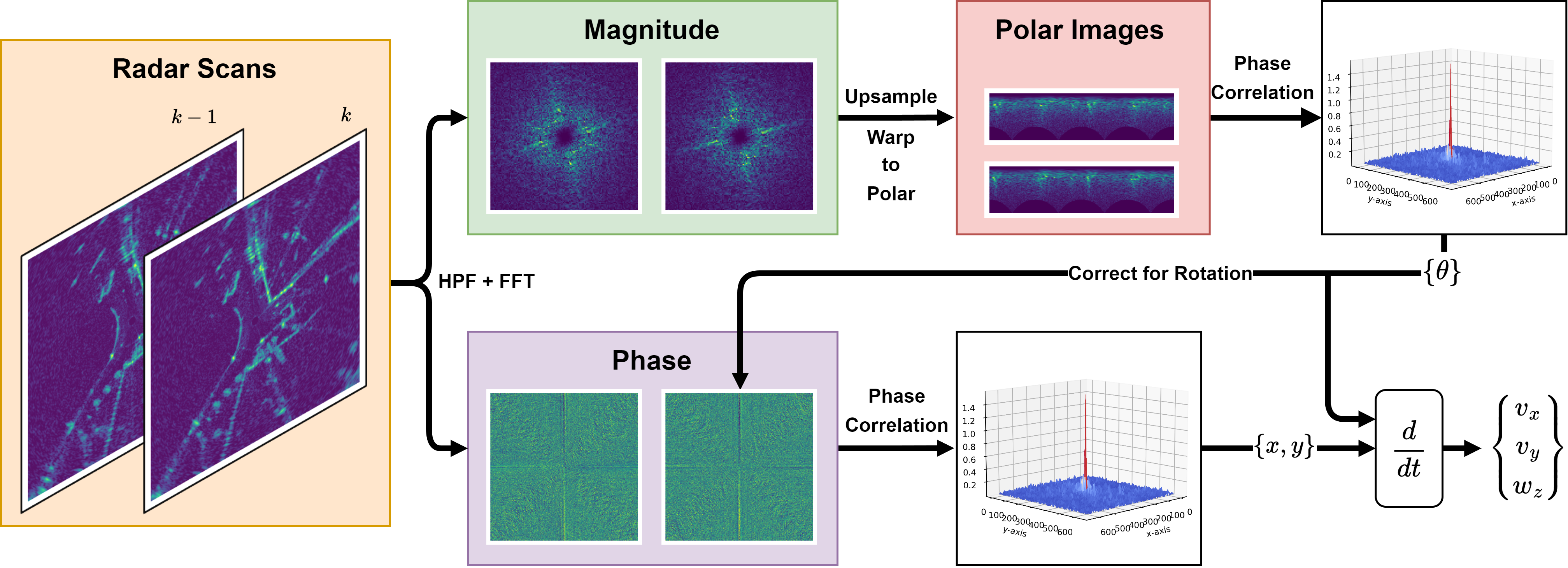}
    \caption{FD-RIO's dense radar odometry pipeline. Translation and rotation between two scans are estimated using a two-step approach: magnitudes of FFT images are warped to polar coordinates, then the phase correlation is calculated to find the rotation, one scan is then corrected for rotation and phase correlation is used again to estimate the translation.}\label{fig:algorithmi}
\end{figure*}
\subsection{State Estimation with Kalman Filter}\label{subseq:kf}
Kalman filter is a classic state estimator that combines measurements with state predictions that are based on a process model. It also incorporates uncertainties associated with both the measurements and predictions. Kalman filter and its variant have been exhaustively discussed in the literature. Interested readers can find many resources on the topic such as \cite{welch1995introduction} and \cite{chen2011kalman}. In our proposed method, we estimate five states only to keep the computational overhead of estimation and fusion minimum. The states to be estimated are the two linear velocities $v_x$ and $v_y$, the angular velocity $w_z$, and the two linear accelerations $\alpha_x$ and $\alpha_y$. Our state vector is defined as:
\begin{equation}
    \label{equ:state_vector}
    x_k = \begin{bmatrix} v_x & v_y & w_z & \alpha_x & \alpha_y \end{bmatrix}
\end{equation}
These states are sufficient for problems formulated in $SE(2)$ similar to our 2D-based scanning radar odometry. The Kalman filter updates the states asynchronously whenever a measurement is taken. IMU measurements go directly into the Kalman filter; however, the radar scans are processed in the phase correlation pipeline (subsection \ref{subseq:pc_ro}) to generate linear and angular velocity estimates which in turn are sent to the Kalman filter as shown in Fig. \ref{fig:top_level}. At every new measurement, regardless of its source, an \textit{update} step is triggered which updates the states based on observations made by the sensors on the environment. Each \textit{update} is followed by a \textit{predict} step that uses a constant acceleration motion model, the assumption of constant acceleration is a reasonable assumption here as changes in the acceleration at a typical IMU frequency (e.g., 100-200Hz) can be considered insignificant.

We follow classical Kalman filter formulation as in equations (\ref{equ:state_prediction}) for states update and (\ref{equ:covariance_prediction}) for  covariance update: 
\begin{equation}
    \label{equ:state_prediction}
    x_{k+1} = F_k x_k + B_k u_k 
\end{equation}
\begin{equation}
    \label{equ:covariance_prediction}
    P_{k+1} = F_k P_k F_k^{T} + Q_k
\end{equation}
where $F_k$ is the state transition matrix, $B_k$ is the control matrix, $u_k$ is the inputs vector, $P_k$ is the states covariance matrix, and $Q_k$ is the process noise. In our implementation, the gyro bias is included as a constant input to the system in $u_k$, and the control matrix $B_k$ maps this value to the angular velocity $w_z$. The \textit{update} step follows equations (\ref{equ:kalman_gain}), (\ref{equ:state_update}), and (\ref{equ:covariance_update}):
\begin{equation}
    \label{equ:kalman_gain}
    K = P_k H_k^T (H_k P_k H_k^T + R_k)^{-1}
\end{equation}
\begin{equation}
    \label{equ:state_update}
    x_k^\prime = x_k + K(z_k - H_k x_k)
\end{equation}
\begin{equation}
    \label{equ:covariance_update}
    P_k^\prime = P_k - K H_k x_k
\end{equation}
where $K$ is the Kalman gain, $H_k$ is the measurement matrix, $R_k$ is the measurements noise, and $z_k$ is the measurement vector. Since we are performing an \textit{update} step from two different sensors, our formulation has two distinct measurement matrices, $H_{k,radar}$ that maps radar odometry estimates to states $v_x$. $v_y$, and $w_z$, and $H_{k,imu}$ that maps the appropriate accelerometer and gyro measurements to states $w_z$, $\alpha_x$, and $\alpha_y$.
\subsection{Phase Correlation and Radar Odometry}\label{subseq:pc_ro}
Our proposed state estimator builds on the work presented in \cite{fft_reddy} and \cite{checchin2010radar} and runs in two steps, one to estimate the rotation angle and another to estimate the translation as illustrated in Fig. \ref{fig:algorithmi}. Radar scans are processed as grayscale images and downsampled to 640x640 pixels to keep computations as efficient as possible. Moreover, our dense method does not require the scans to be compensated for motion distortion and we do not implement a keyframe selection strategy. Assuming two consecutive scans $I_{k-1}(x,y)$ and $I_k(x,y)$ with shifts in both dimensions $(\Delta x,\Delta y)$ such that:
\begin{equation}
    \label{equ:shift}
    I_k(x,y) = I_{k-1}(x-\Delta x, y - \Delta y)
\end{equation}
The phase correlation can be calculated using:
\begin{equation} \label{equ:pc}
    PC = \frac{\hat{I}_k \hat{I}_{k-1}^*}{\lvert\hat{I}_k \hat{I}_{k-1}^*\rvert}
\end{equation}
where $\hat{I}$ is the image in frequency domain after applying Fourier transform such that $\hat{I}_k(u,v)=\mathcal{F}(I_k(x,y))$, and ($^*$) is the complex conjugate. The two shifts $(\Delta x,\Delta y)$ can then be calculated using:
\begin{equation} \label{equ:deltas}
    \Delta x, \Delta y = \underset{x,y}{\arg\max}\{ \mathcal{F}^{-1}(PC)\}
\end{equation}
where the lateral translation $\Delta y$ is expected to be very small for Ackermann geometry -based drive.
\subsubsection{Rotation Estimation}
The rotation angle $\theta$ between two consecutive radar scans $I_{k-1}(x,y)$ and $I_k(x,y)$ is retrieved by first passing the two images through a High Pass Filter (HPF), this is done to strengthen the edges and improve the correlation accuracy. Magnitude images of the Fourier transform are obtained for both images by applying the Fast Fourier Transform (FFT), based on the Fourier shift theorem; the two magnitude images will not carry translation differences between the original images, but they will be, however, rotated by the angle $\theta$ which can be extracted by warping both images to polar coordinates and calculating the phase correlation using equations (\ref{equ:pc}) and (\ref{equ:deltas}). In order to improve the precision of the angle estimates we upsample the polar images using linear interpolation. In our implementation, an upsampling factor of 4 was found sufficient as will be discussed in Table \ref{tab:mulran_results} and Table \ref{tab:boreas_results} in Section \ref{sec:results}. Higher upsampling factors are also possible and would further improve the precision but they come at a computational cost that we are trying to keep minimum. Note that the second shift pertains the scale change between the two images which is irrelevant to our radar odometry algorithm.

\subsubsection{Translation Estimation}

After calculating an estimate of the rotation angle $\theta$, one of the two scans is corrected for rotation in the spatial domain so that the two scans now only differ by translation. Next, the same phase correlation approach, equations (\ref{equ:pc}) and (\ref{equ:deltas}), is used to estimate the shifts in the \textit{x} and \textit{y} -axes in the units of pixels. Those values are mapped to actual translation values in meters based on the resolution of the radar images. Finally, based on those translation and rotation values, the linear and angular velocities ($v_x$, $v_y$, and $w_z$) are calculated using the time elapsed between the two radar scans.

\section{Experimental Results}\label{sec:results}
We tested our method on two publicly available datasets: \textit{MulRan} dataset\cite{mulran_dataset} and \textit{Boreas} dataset\cite{boreas_dataset}. The results are reported here using KITTI evaluation metrics\cite{KITTI}: The Average Translational Error and Average Rotational Error which are the most commonly used metrics for odometry evaluation. These metrics are calculated by averaging the relative translation error and relative rotation error over sub-trajectories of lengths (100, 200, 300, 400, 500, 600, 700, and 800) meters. The resulting values are reported as (translational error [\%] / rotational error [deg/100m]). We used the open-source toolkit developed by Zhan \textit{et al.}\cite{what_should_be_learnt} to run all our evaluations. In our tests, we clip IMU readings before the first radar scan and after the last radar scan as our method relies on the fusion of both sensors. The first pose of the odometry estimation was initialized from ground truth poses, and the trajectories were plotted without any alignment techniques (e.g., Umeyama alignment). Since the estimated poses in our asynchronous fusion approach and the ground truth poses do not always map one to one, we filled the gaps using linear interpolation for positions and Spherical Linear Interpolation (SLERP) for orientation. The only parameter that had to be tuned between the two datasets is the gyro bias as it depends on the sensor used during collection. Finally, tests were not carried out on the popular \textit{Oxford} dataset\cite{oxford_radar_dataset} due to the unavailability of raw IMU measurements in the published data.

\begin{figure*}[!t]
    \centering
    \subfloat[]{\includegraphics[width=3.2in]{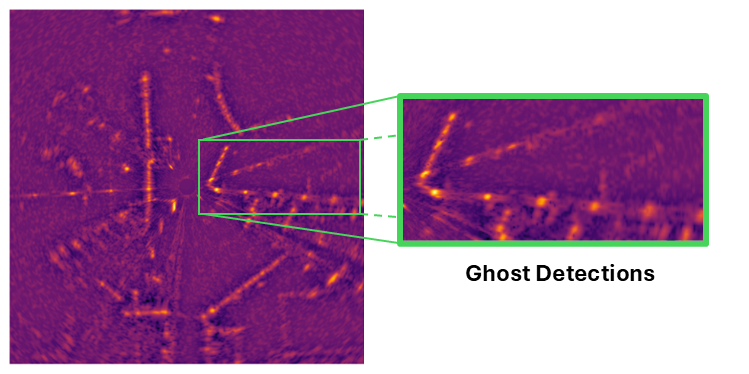}\label{fig:example_a}}\hfil
    \subfloat[]{\includegraphics[width=3.2in]{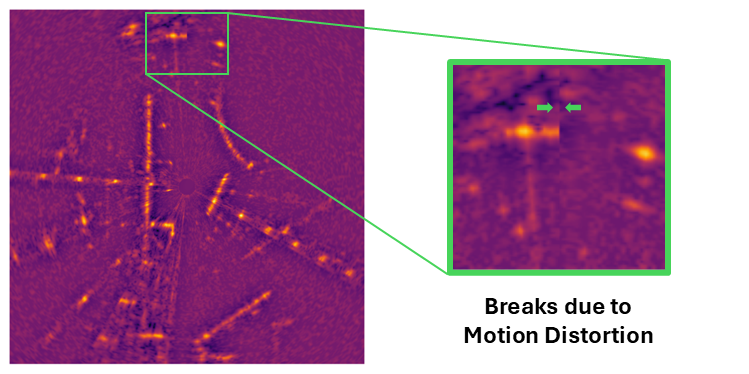}\label{fig:example_b}}\hfil
    \subfloat[]{\includegraphics[width=3.2in]{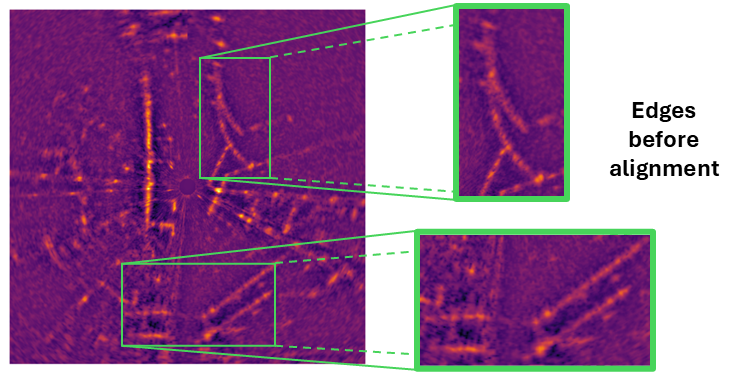}\label{fig:example_c}}\hfil
    \subfloat[]{\includegraphics[width=3.2in]{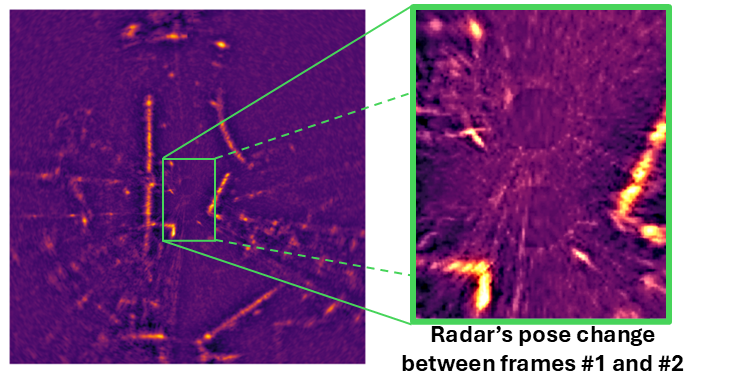}\label{fig:example_d}}\hfil
    \caption{Two radar scans taken from the \textit{Boreas} sequence \textit{2021-04-08-12-44}. This example illustrates the registration performed by FD-RIO's odometry pipeline despite the noticeable motion distortion and ghost detections. It is important to note that the bright areas are not detections/features; FD-RIO is a dense method that does not rely on detections. The bright areas are stronger returns displayed as pixels with higher intensities for better visualization. (a) Sample frame \#1. (b) Sample frame \#2. (c) Both frames overlaid highlighting the pose change. (d) Both frames after registration.}
    \label{fig:alginment_example}
\end{figure*}

\begin{figure*}[!t]
    \centering
    \subfloat[]{\includegraphics[height=1.9in, width=2.0in]{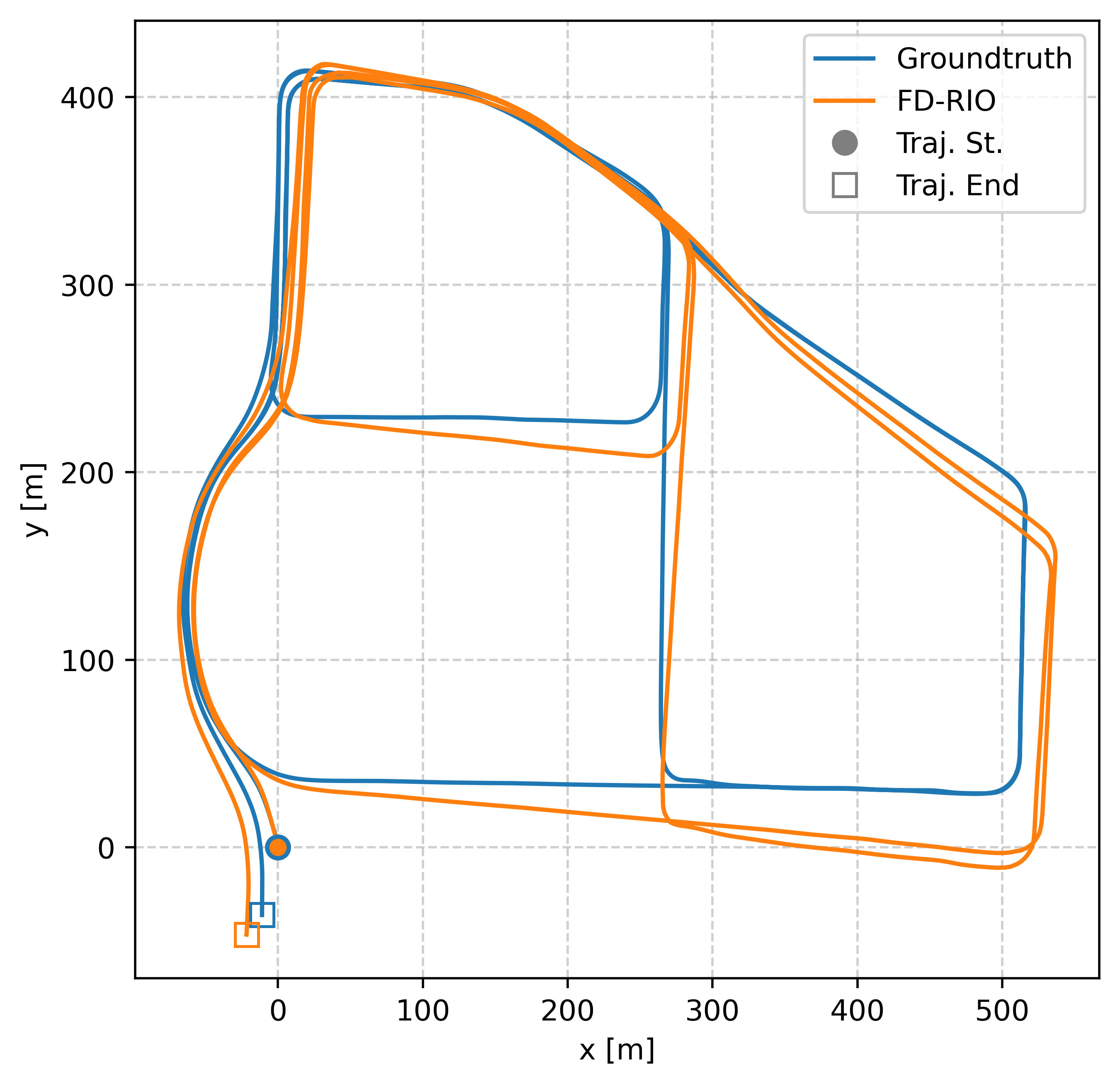}\label{fig:mulran_f1}}\hfil
    \subfloat[]{\includegraphics[height=1.9in, width=2.0in]{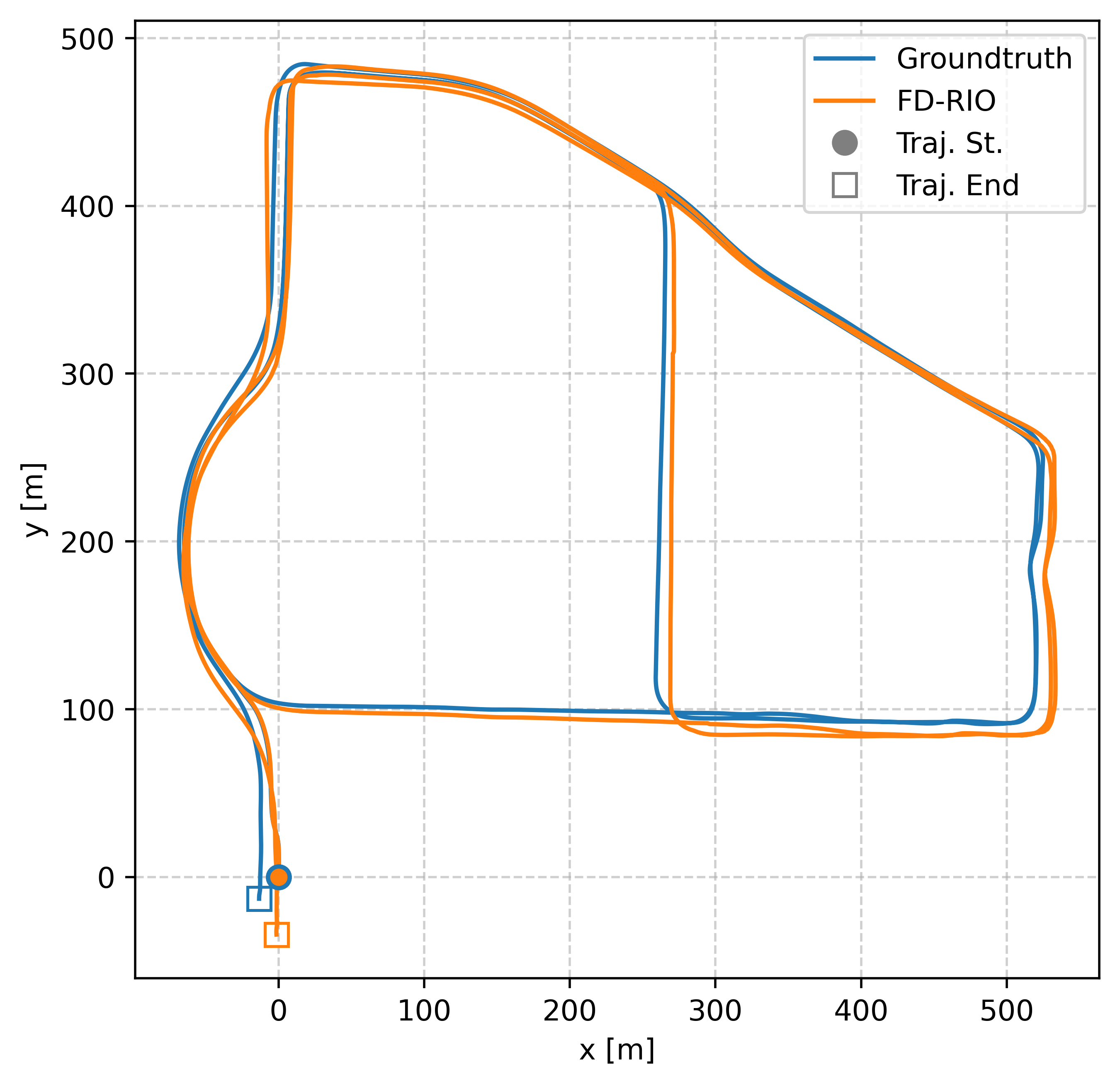}\label{fig:mulran_f2}}\hfil
    \subfloat[]{\includegraphics[height=1.9in, width=2.0in]{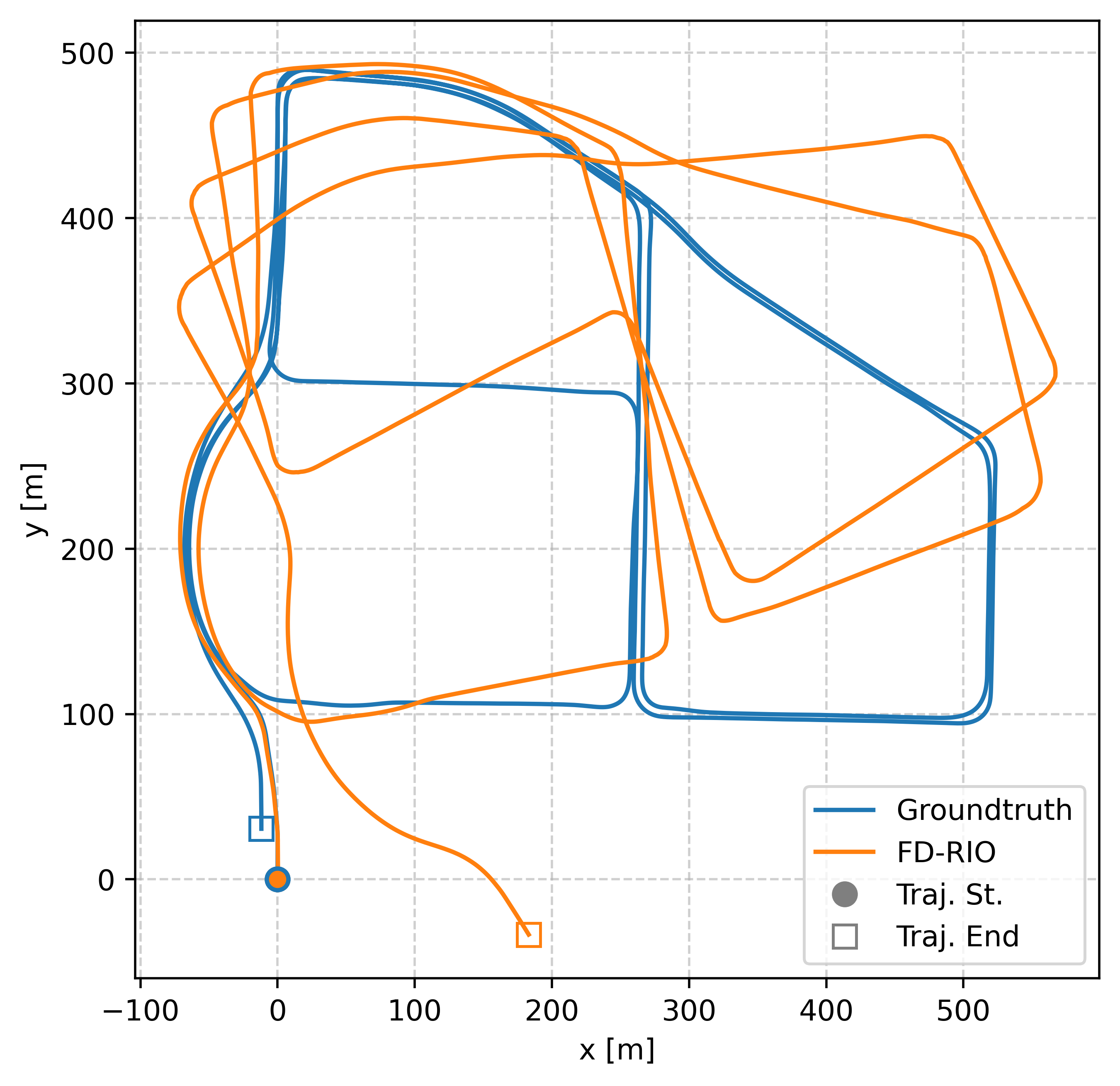}\label{fig:mulran_f3}}\hfil
    \subfloat[]{\includegraphics[height=1.9in, width=2.0in]{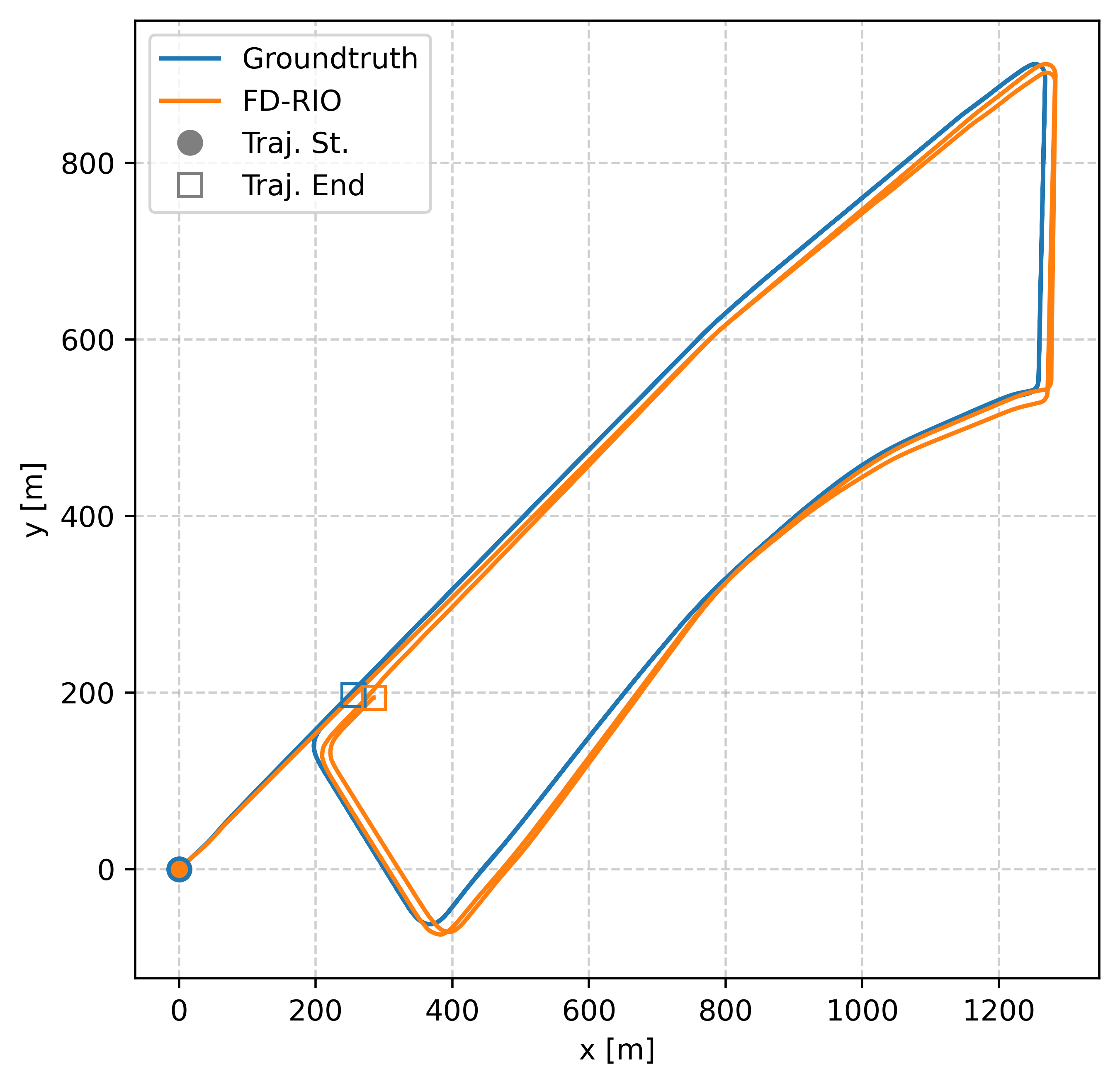}\label{fig:mulran_f4}}\hfil
    \subfloat[]{\includegraphics[height=1.9in, width=2.0in]{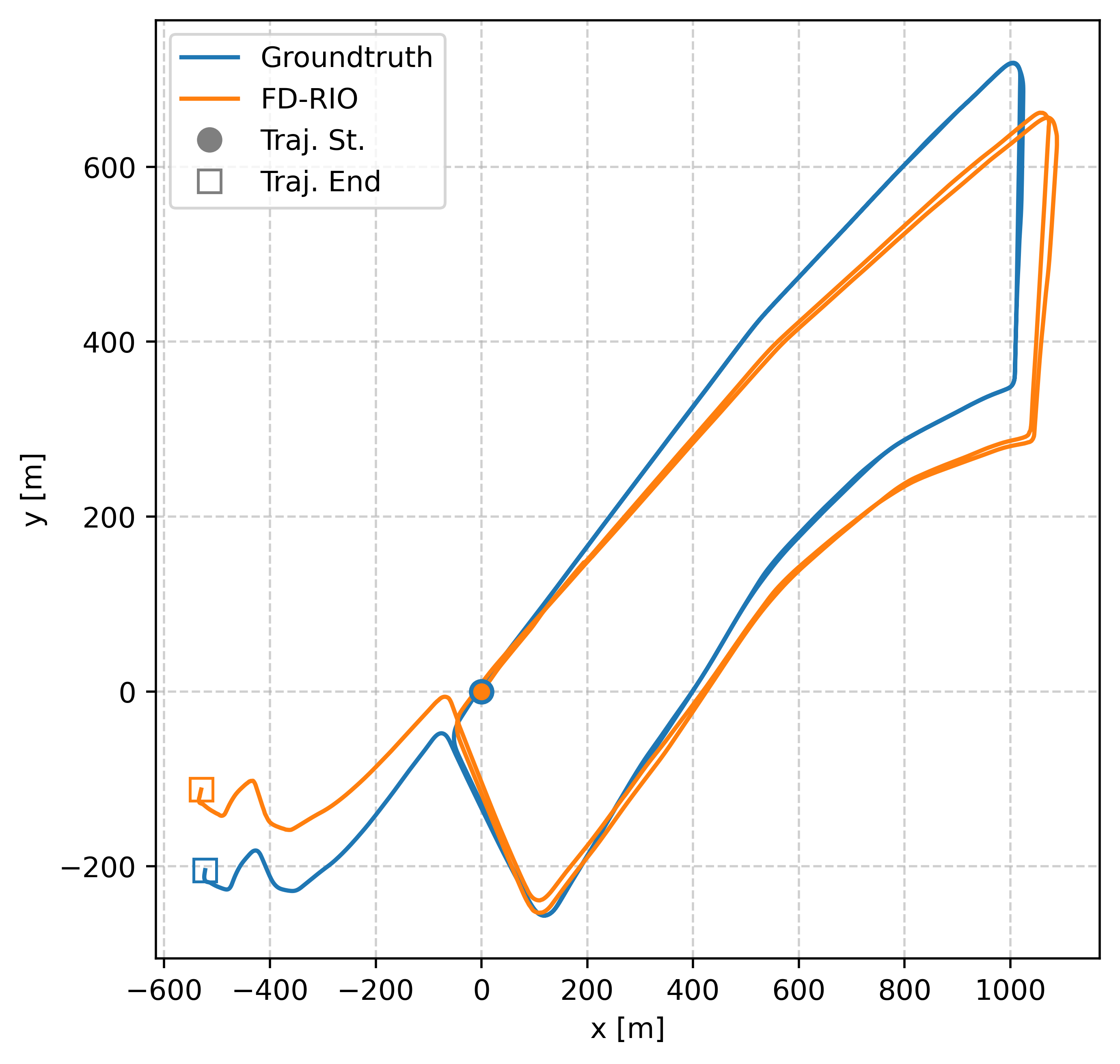}\label{fig:mulran_f5}}\hfil
    \subfloat[]{\includegraphics[height=1.9in, width=2.0in]{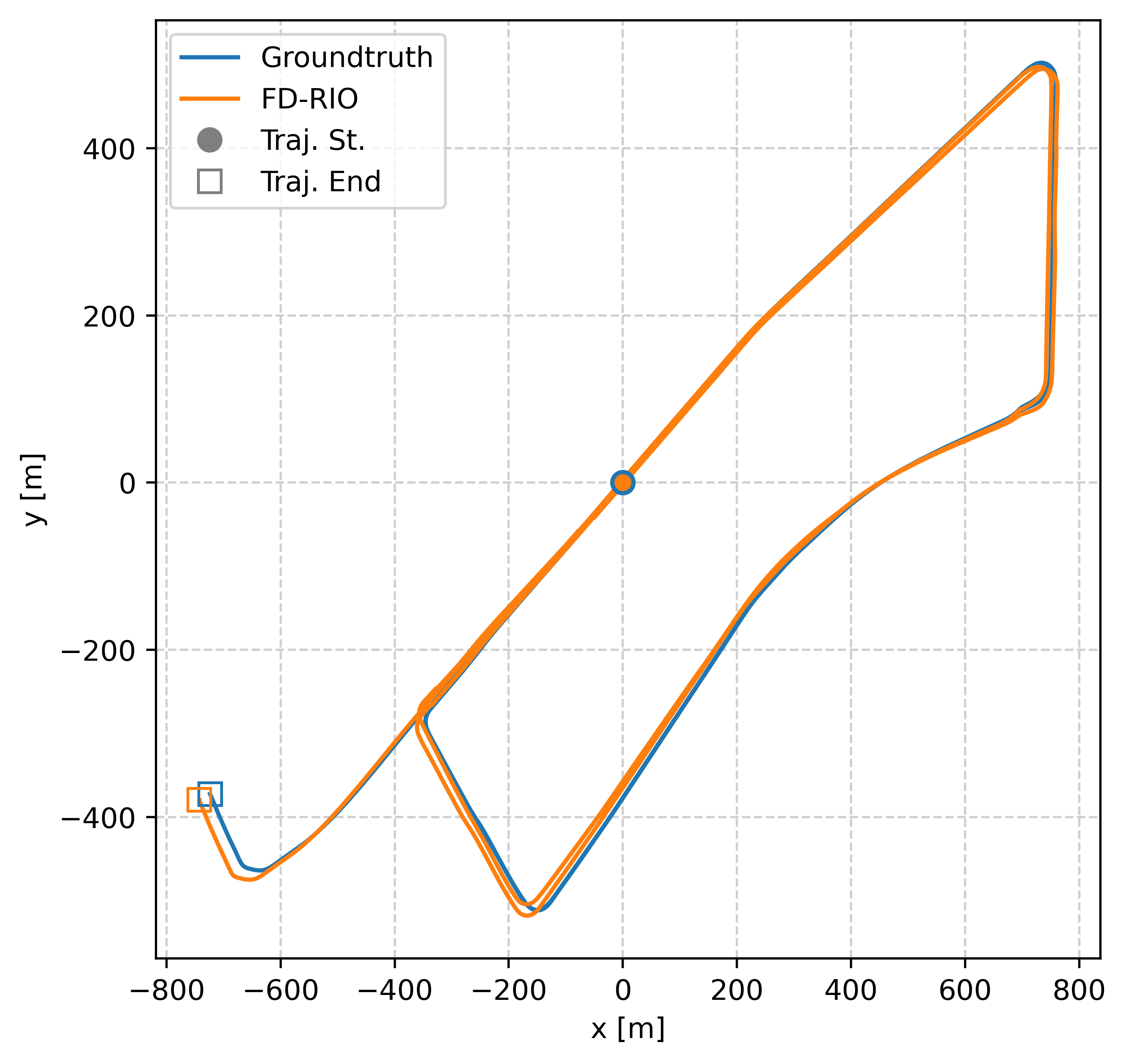}\label{fig:mulran_f6}}\hfil
    \subfloat[]{\includegraphics[height=1.9in, width=2.0in]{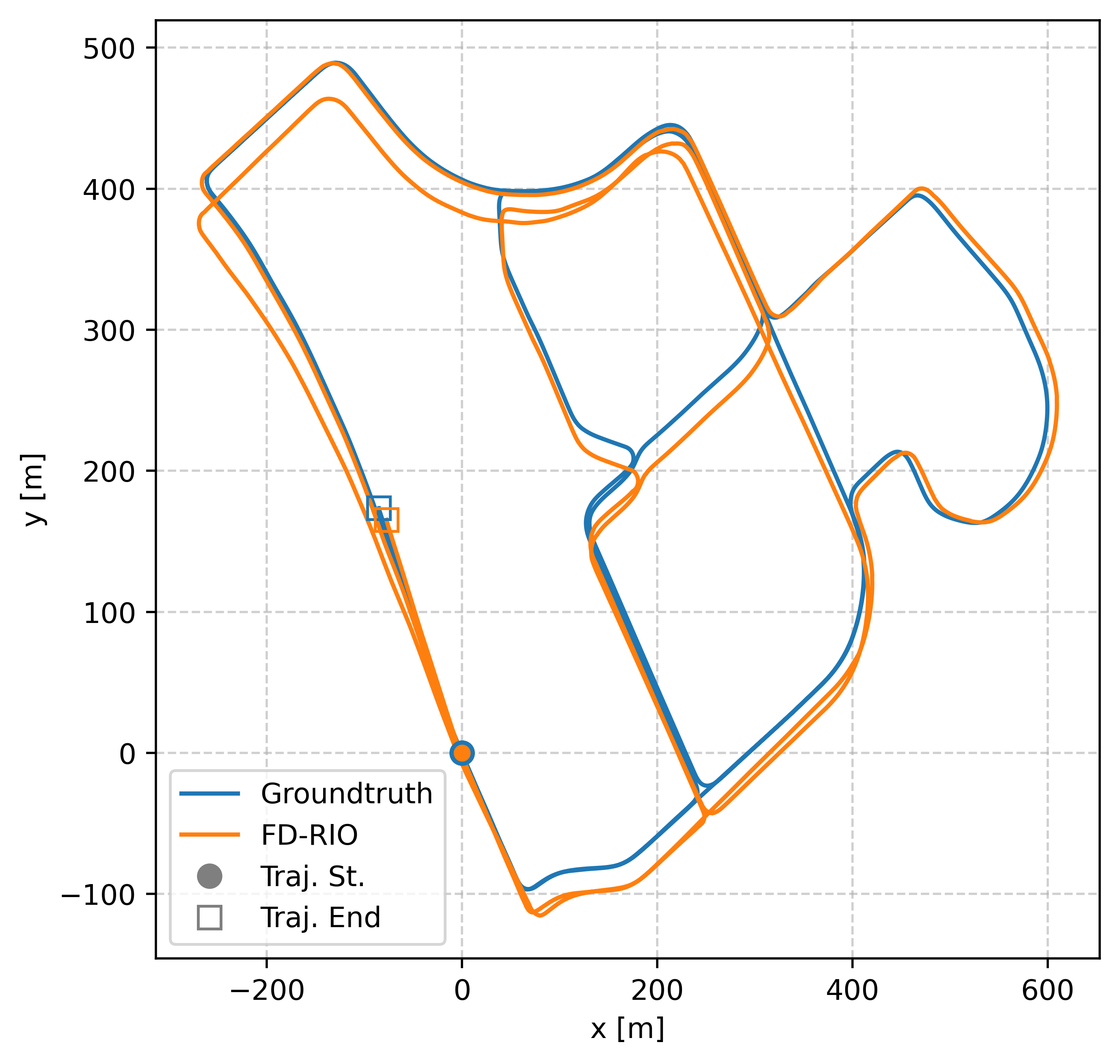}\label{fig:mulran_f7}}\hfil
    \subfloat[]{\includegraphics[height=1.9in, width=2.0in]{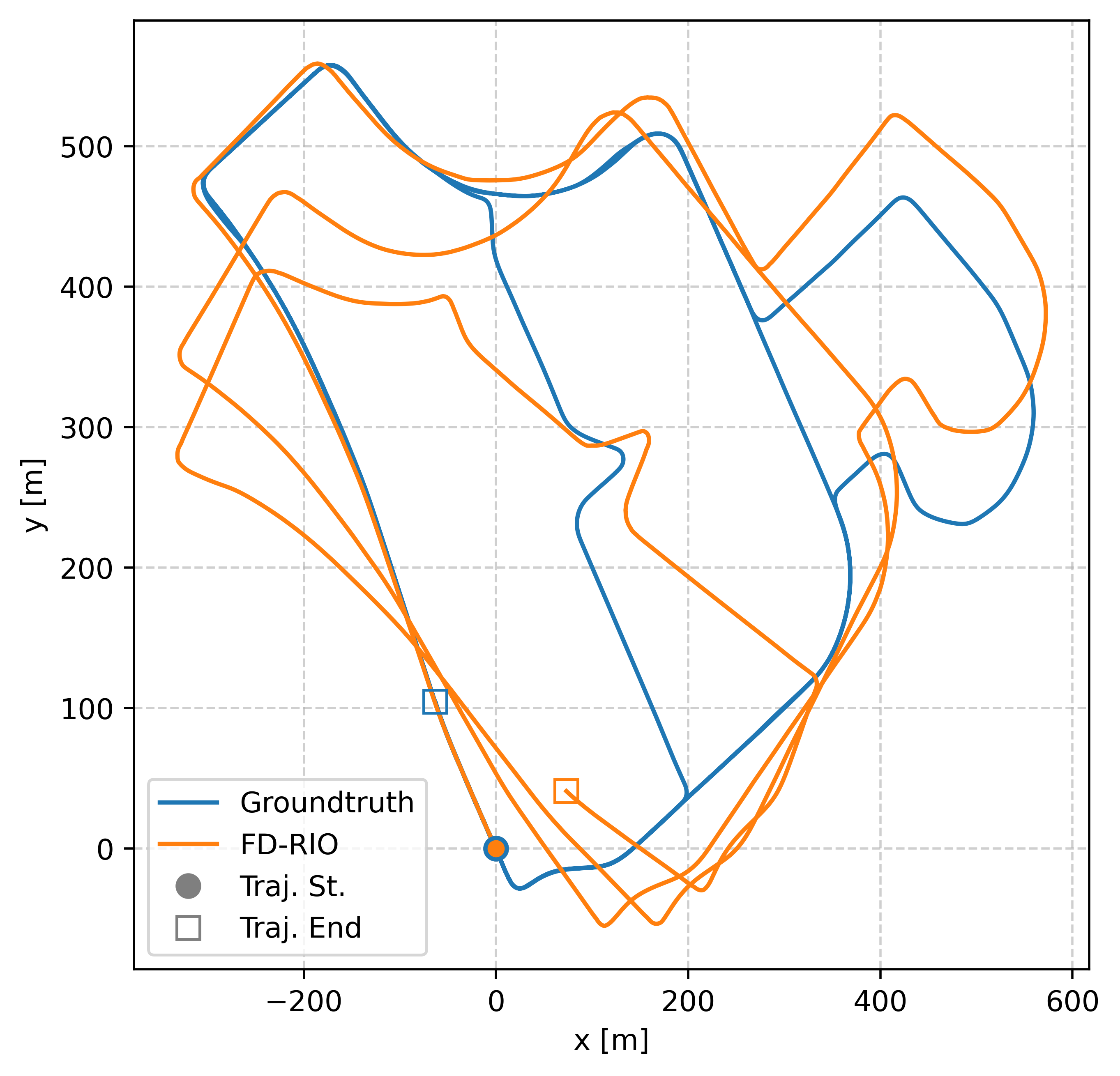}\label{fig:mulran_f8}}\hfil
    \subfloat[]{\includegraphics[height=1.9in, width=2.0in]{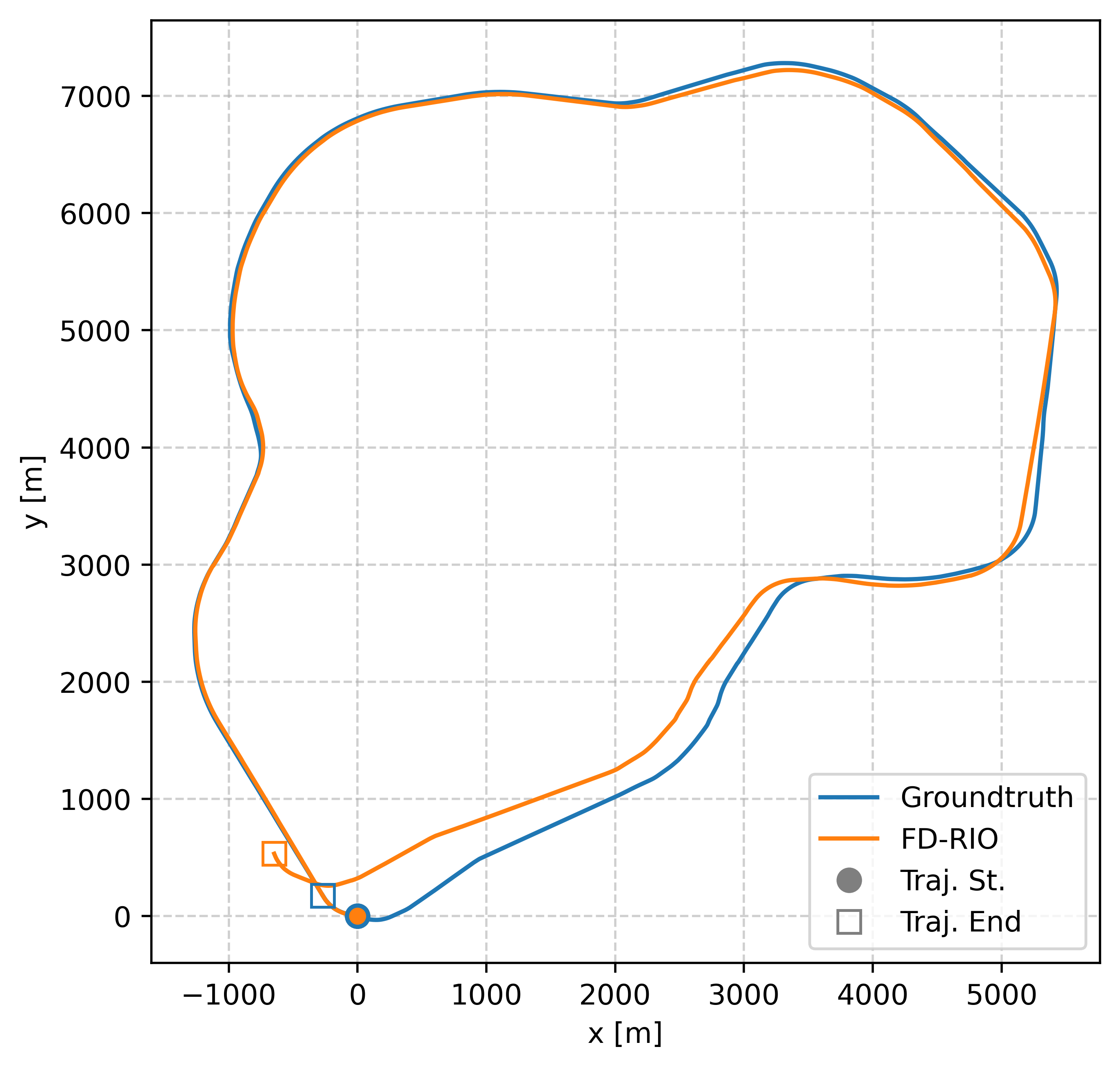}\label{fig:mulran_f9}}\hfil
    \subfloat[]{\includegraphics[height=1.9in, width=2.0in]{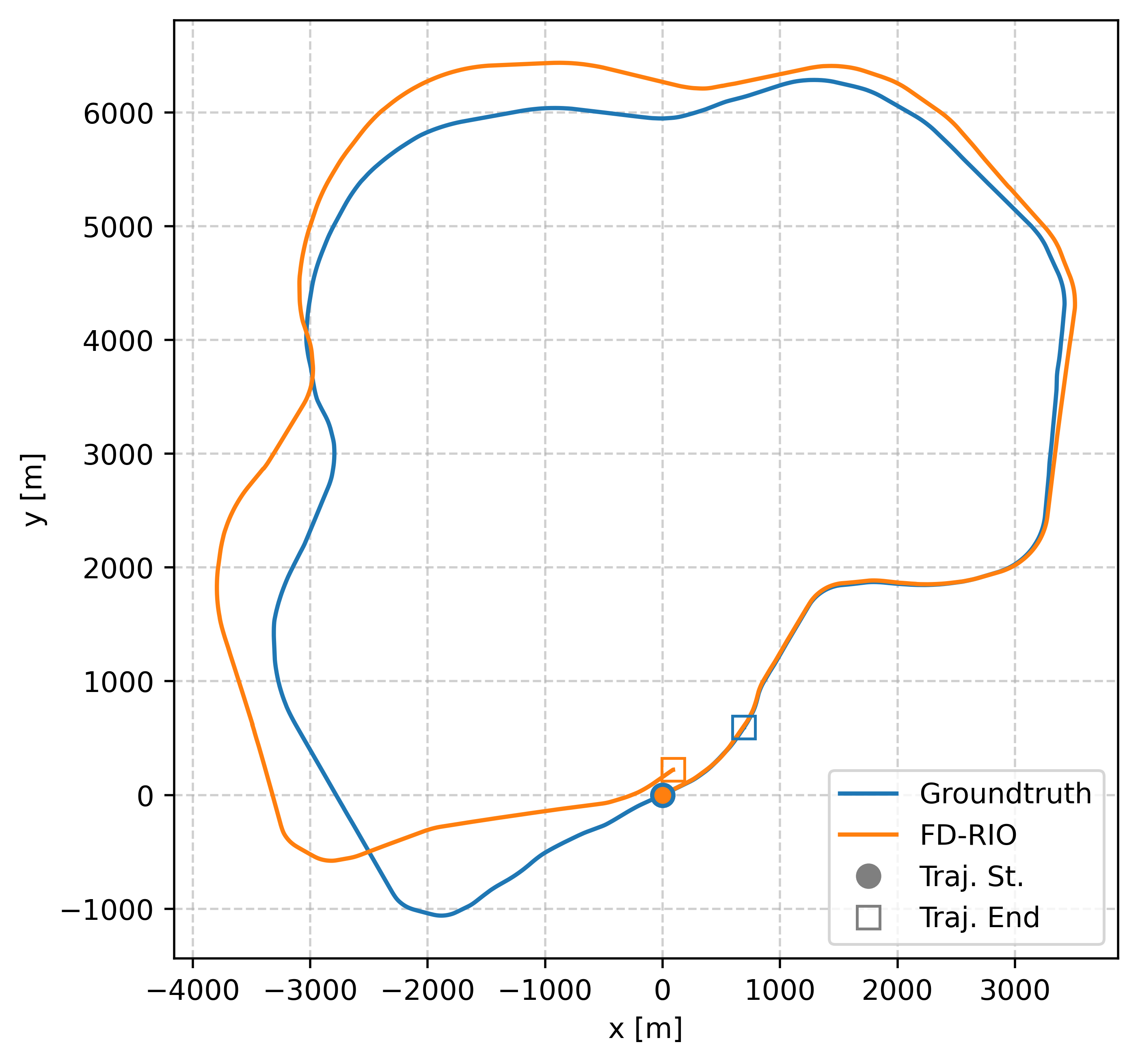}\label{fig:mulran_f10}}\hfil
    \caption{Test results using \textit{MulRan} dataset. Estimated trajectories by FD-RIO vs ground truth. Sequences shown are: (a) DCC01. (b) DCC02. (c) DCC03. (d) Riverside01. (e) Riverside02. (f) Riverside03. (g) KAIST02. (h) KAIST03. (i) Sejong02. (j) Sejong03}
    \label{fig:mulran_trajectories}
\end{figure*}

\begin{figure*}[!t]
    \centering
    \subfloat[]{\includegraphics[width=1.7in]{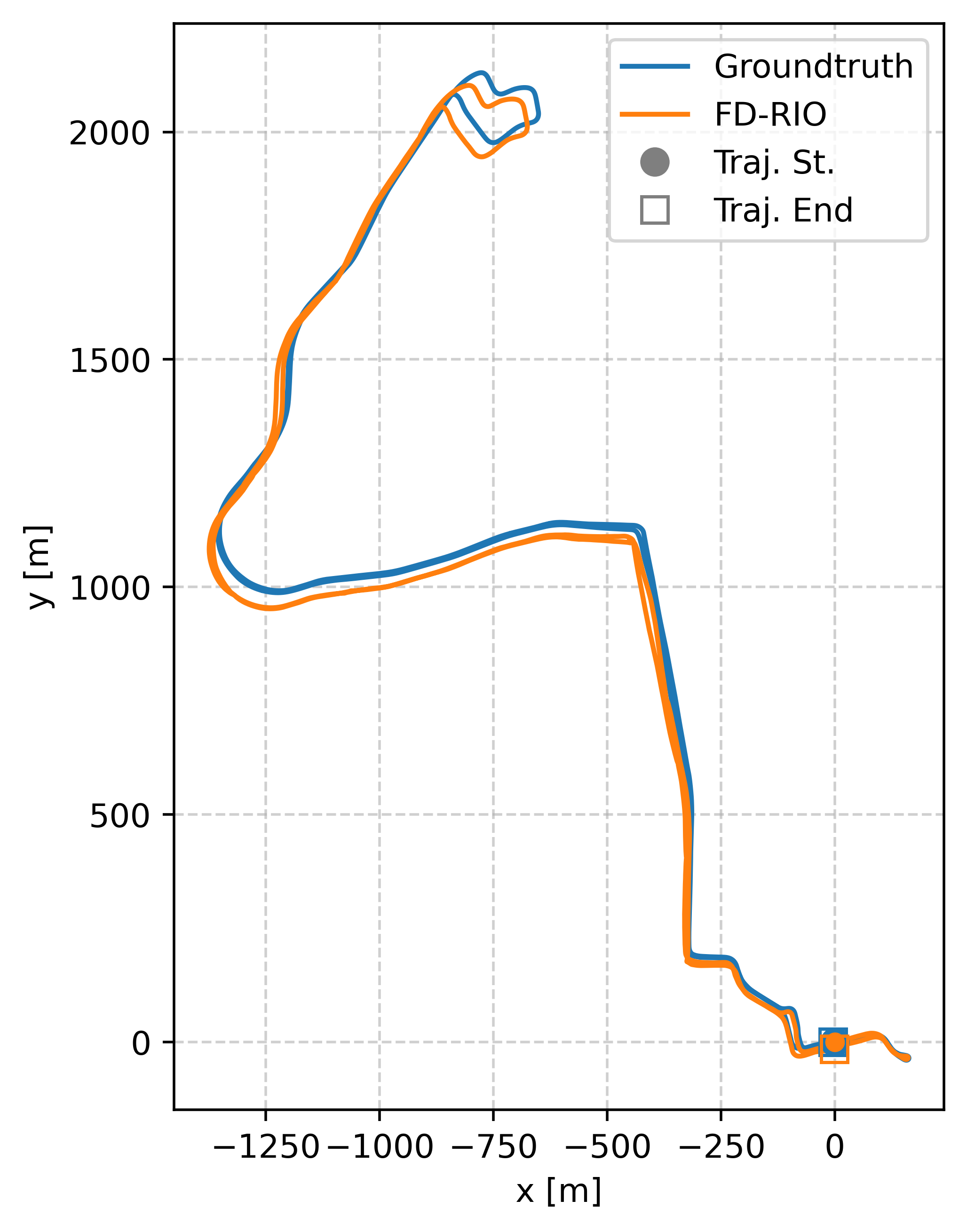}\label{fig:boreas_f1}}\hfil
    \subfloat[]{\includegraphics[width=1.7in]{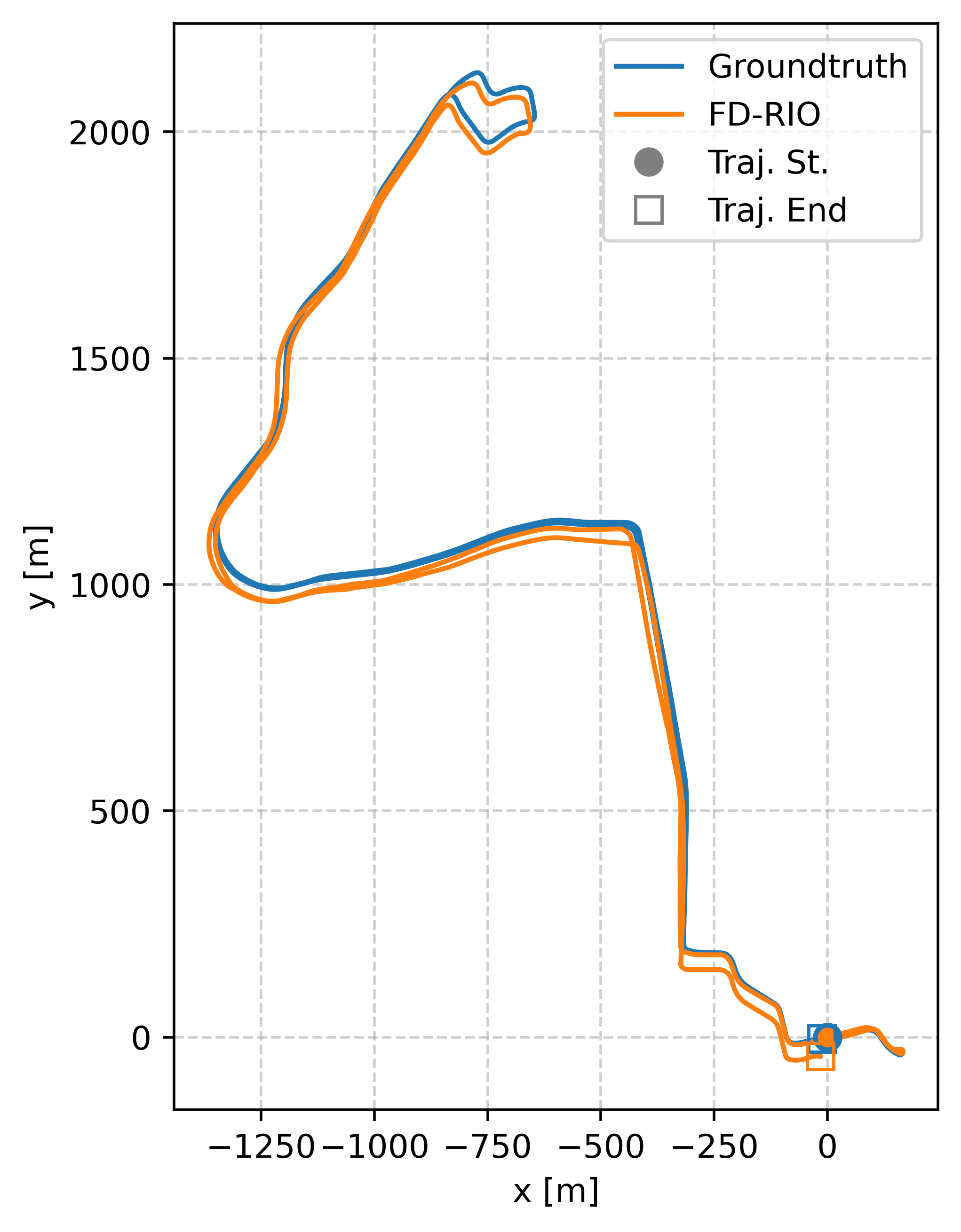}\label{fig:boreas_f2}}\hfil
    \subfloat[]{\includegraphics[width=1.7in]{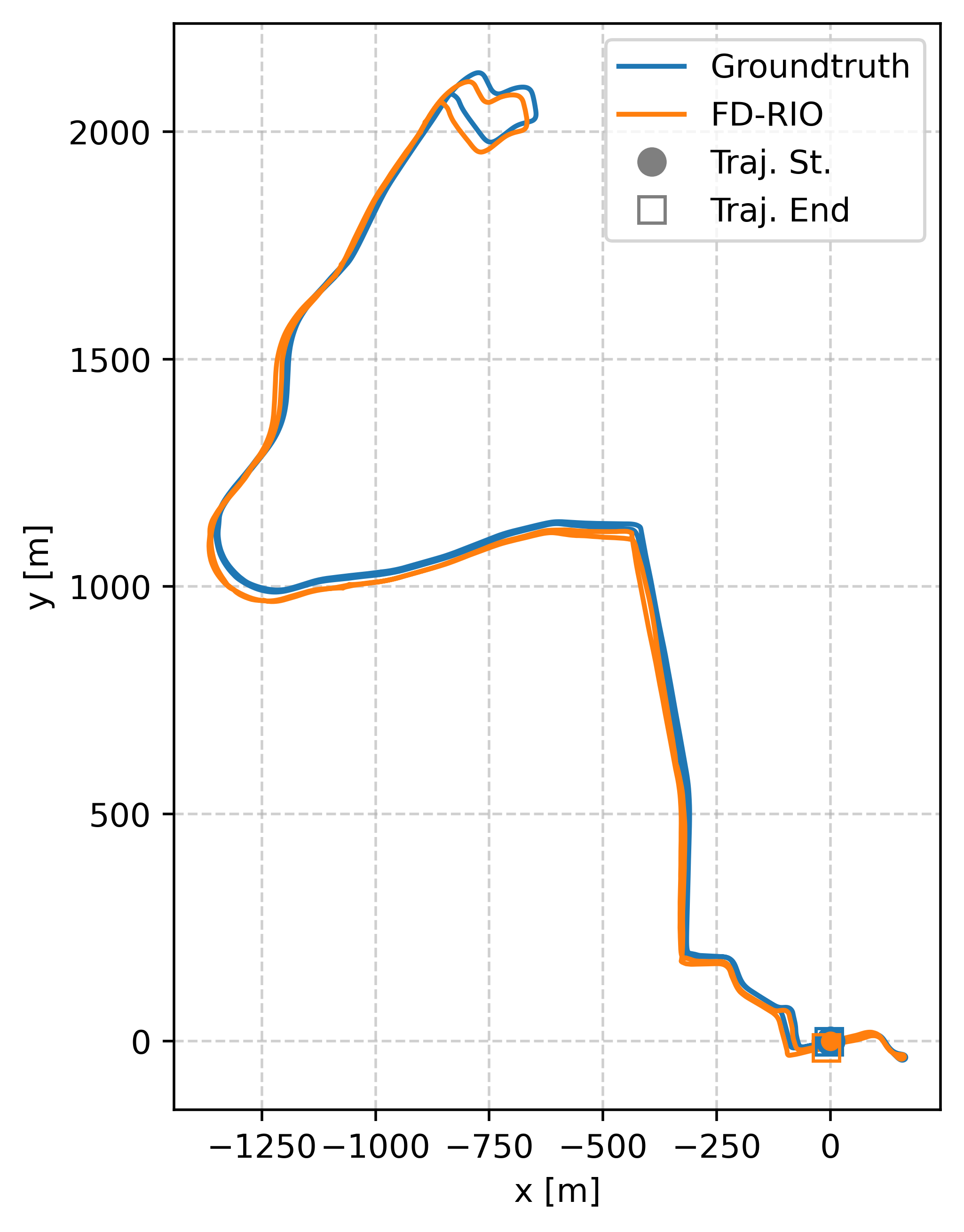}\label{fig:boreas_f3}}\hfil
    \subfloat[]{\includegraphics[width=1.7in]{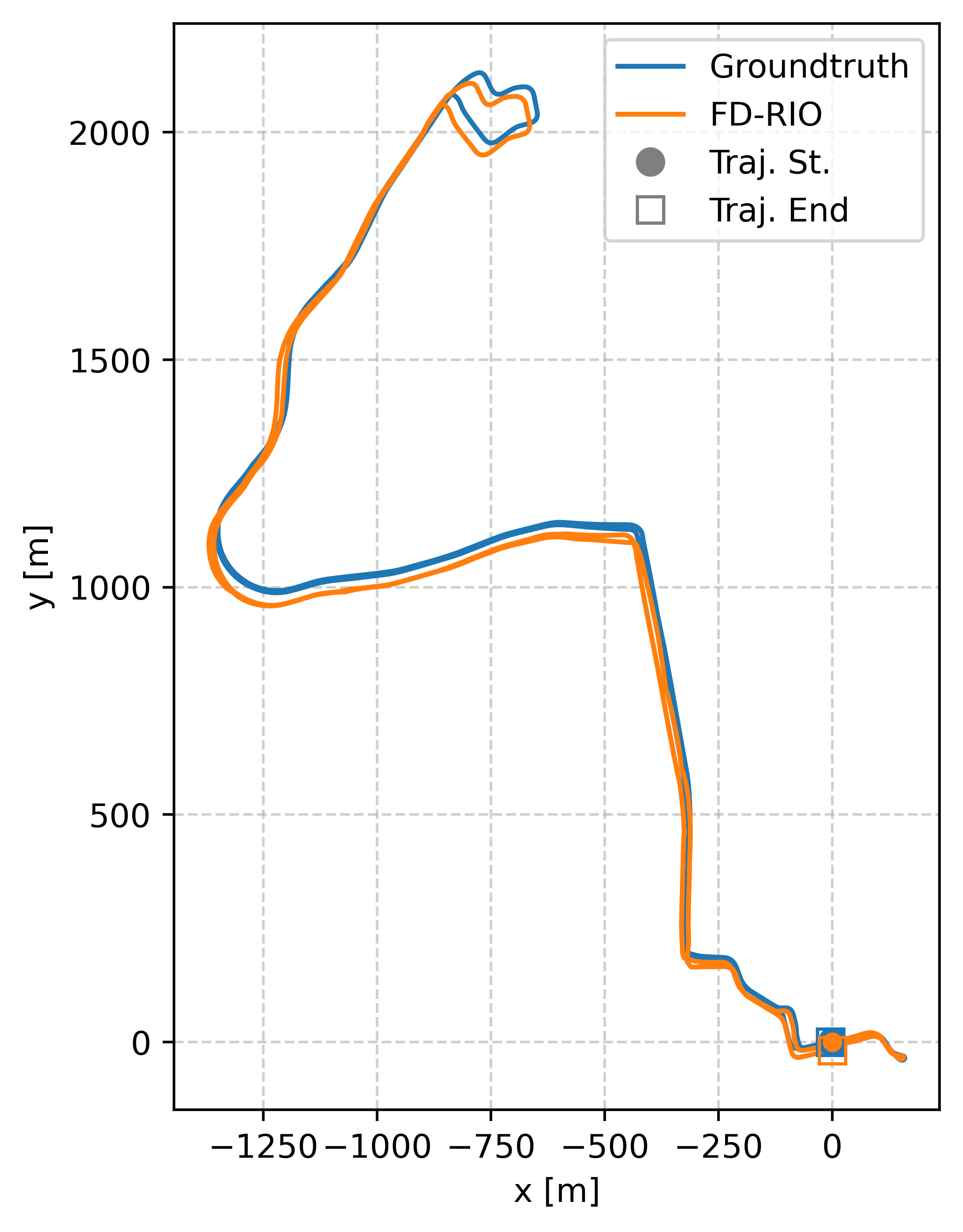}\label{fig:boreas_f4}}\hfil
    \caption{Test results using \textit{Boreas} dataset. Estimated trajectories by FD-RIO vs ground truth. Sequences shown are: (a) 2020-11-26-13-58. (b) 2021-01-26-11-22. (c) 2021-04-08-12-44. (d) 2021-04-29-15-55.}
    \label{fig:boreas_trajectories}
\end{figure*}

\subsection{Tests on \textit{MulRan} Dataset}\label{subsec:mulran_tests}
Our main tests were carried out using the \textit{MulRan} dataset \cite{mulran_dataset} which was collected using a lidar sensor and a scanning radar (Navtech, CIR204-H). The dataset has four trajectories (i.e., \textit{DCC}, \textit{KAIST}, \textit{Riverside} and \textit{Sejong}), it was collected for place recognition research and there are multiple runs for each trajectory. \textit{Sejong} route is significantly longer than the others; therefore, it is substantially more challenging for odometry algorithms. Our test results are summarized in Table \ref{tab:mulran_results} along with results for other state-of-the-art methods. The table clearly shows the viability of FD-RIO; it exceeds the state-of-the-art in four sequences (i.e., \textit{KAIST02}, \textit{Riverside03}, \textit{Sejong02}, and \textit{Sejong03}) both in translational and rotational error, and is very competitive on all other sequences. Furthermore, FD-RIO offers substantial improvement on what have been previously reported in the literature for the longer sequences \textit{Sejong02} and \textit{Sejong03}. Fig. \ref{fig:mulran_trajectories} shows the estimation vs ground truth for all sequences reported in Table \ref{tab:mulran_results}. 

\begin{table*}
    \begin{center}
    
    \caption{Evalution results on ten sequences from \textit{MulRan} Dataset}\label{tab:mulran_results}
    \vspace{-8pt}
    \resizebox{\linewidth}{!}{%
    \begin{tabular}{l|c|c|cccccccccc|c}
        \hline
              &       &     & \multicolumn{10}{c|}{\textbf{Sequence}} &                \\
        \textbf{Method} & \textbf{Evaluation} & \textbf{Type} & \textbf{DCC01} &  \textbf{DCC02} &  \textbf{DCC03} & \textbf{KAIST02} & \textbf{KAIST03} & \textbf{Riverside01} & \textbf{Riverside02} & \textbf{Riverside03} & \textbf{Sejong02} & \textbf{Sejong03} & Mean \\ \hline \hline

        ORORA\textsuperscript{\textdagger}\cite{lim2023orora}&\cite{lim2023orora}        & S &   3.18/0.66   &   2.38/0.57   &   2.77/0.79   &   3.12/0.78   &   2.53/0.57   &   3.51/0.76   &   3.31/0.76   &   2.79/0.64   &   3.76/0.67   &   5.07/0.86   &   -         \\
        RadarSLAM (odometry) \cite{hong2022radarslam}         &\cite{hong2022radarslam}   & S &   2.70/0.50   &   1.90/0.40   &   1.64/0.40   &   2.07/0.60   &   1.99/0.50   &   2.04/0.50   &   1.51/0.50   &   1.71/0.50   &   -           &   -           &   1.97/0.50 \\
        CFEAR-3 \cite{adolfsson2023cfear}                     &\cite{adolfsson2023cfear}  & S &   2.28/0.54   &   1.49/0.46   &   1.47/0.48   &   1.62/0.66   &   1.73/0.78   &   1.59/0.63   &   1.39/0.51   &   1.41/0.40   &   -           &   -           &   1.62/0.57 \\ 
        CFEAR-3-s50 \cite{adolfsson2023cfear}                 &\cite{adolfsson2023cfear}  & S &   2.09/0.55   &\e{1.38}/0.47  &\e{1.26}/0.47  &   1.51/0.63   &   1.59/0.75   &   1.62/0.62   &\e{1.35}/0.52  &   1.19/0.37   &   -           &   -           &   1.50/0.56 \\
        MC-RANSAC\textsuperscript{\textdaggerdbl}\cite{burnett2021mcransac}&\cite{lim2023orora}        & S &   4.49/1.12   &   3.64/1.00   &   4.37/1.39   &   6.32/1.68   &   4.42/1.12   &   5.92/1.43   &   8.38/2.08   &   6.61/1.88   &   8.83/2.32   &   12.01/2.46   &   -         \\
        SDRO \cite{zhang2023sdr}                             &\cite{zhang2023sdr}        & S &\e{1.55}/0.35  &   1.53/0.33   & 1.60/\e{0.30} &   1.61/0.35   & 1.59/\e{0.32} & 1.61/\e{0.26} & 1.59/\e{0.27} &   1.62/0.29   &   -           &   -           &   1.59/0.31 \\ \hline
        $R^3O$ \cite{r3o}                                    &\cite{r3o}                 & D &   2.39/0.43   &   1.40/0.34   &   1.48/0.41   &   1.55/0.53   &\e{1.53}/0.50  &\e{1.34}/0.39  &   1.98/0.53   &   1.81/0.57   &   -           &   -           &   1.70/0.48 \\
        FD-RIO (Ours)                                        &    -                      & D & 2.13/\e{0.21} & 1.49/\e{0.23} &   2.30/0.69   &\e{1.16}/\e{0.24} &   2.02/0.54&   1.50/0.38   &   1.46/0.31   &\e{1.08}/\e{0.13}&\e{1.44}/\e{0.26}&\e{2.21}/\e{0.48}&   1.68/0.35  \\ \hline

            \multicolumn{14}{l}{Standard KITTI evaluation metrics are used; translational error [\%] / rotational error [deg/100m]. (-) indicates results that are not available or not applicable.}\\
            \multicolumn{14}{l}{(S) stands for Sparse method, (D) is for Dense method. Methods not tested on \textit{Sejong02} and \textit{Sejong03} receive an advantage on the mean as these sequences are usually more challenging.}\\
            \multicolumn{14}{l}{$\dagger$ based on feature detection by \cite{cen2018precise}. $\ddagger$ based on feature detection by \cite{cen2018precise} and includes motion and Doppler distortion compensation.}\\
            \end{tabular}%
            }
\end{center}
\end{table*}

\subsection{Tests on \textit{Boreas} Dataset}\label{subsec:boreas_tests}
\textit{Boreas} dataset \cite{boreas_dataset} is a publicly available multimodal, multi-seasonal, autonomous driving dataset. It contains scanning radar data (Navtech, CIR304-H) in addition to lidar, camera, and IMU. We leveraged its great seasonal diversity by selecting four sequences that represent various driving conditions (i.e., overcast, snow, sun, and rain). The sequences were picked from the training split where the ground truth data are available and evaluating estimated trajectories is possible. Unfortunately, the only comparable results we found in the literature is what is reported in \cite{burnett2021radar} where only one of the sequences was tested as shown in Table \ref{tab:boreas_results}. Nevertheless, we report our tests on \textit{Boreas} dataset here to verify that FD-RIO can generalize over other datasets and that it is not tailored for the environment or sensor setup of \textit{MulRan} dataset. Table \ref{tab:boreas_results} shows very promising results where FD-RIO is on par with the state-of-the-art in terms of translational error and boasts significant improvement on the rotational error. Fig. \ref{fig:boreas_trajectories} shows the four sequences plotted against the ground truth for comparison. Finally, our tests showed that due to the relatively slower driving speed of \textit{Boreas} dataset compared to \textit{MulRan}, running our method on every other radar scan improves the overall accuracy; this is attributed to having better phase correlation results when the offsets between the scans are more distinguishable, it also suggests that our approach would perform well on higher driving speeds.

Fig. \ref{fig:alginment_example} shows an example taken from the \textit{Boreas} sequence \textit{2021-04-08-12-44} where the pose changes between two scans, as estimated by our proposed method, is used to align them to better visualize its effectiveness. Fig. \ref{fig:example_a} shows a radar scan with visible ghost detections. These detections are typically preprocessed and filtered out as they can have a degrading effect on the ego motion estimation. Similarly, Fig. \ref{fig:example_b} shows another radar scan with slight motion distortion at the top where the beginning and end of two consecutive radar sweeps meet. This sort of artifacts may also require preprocessing. Fig. \ref{fig:example_c} shows the two scans superimposed highlighting the differences. Finally, Fig. \ref{fig:example_d} shows the same two scans after alignment where important road characteristics (i.e., lines and curves in this particular example) are overlapping. Since our method relies mainly on the maximum phase correlation between two scans; this example demonstrates that satisfactory state estimation without the need for expensive preprocessing is attainable.

\begin{table*}
    \begin{center}
    \caption{Evalution results on four sequences from \textit{Boreas} Dataset}\label{tab:boreas_results}
    \resizebox{\linewidth}{!}{%
    \begin{tabular}{l|c|c|cccc|c}
        \hline
              &     &       & \multicolumn{4}{c|}{\textbf{Sequence}} &                \\
        \textbf{Method} & \textbf{Evaluation} &  \textbf{Type} & \textbf{2020-11-26-13-58} & \textbf{2021-01-26-11-22}          & \textbf{2021-04-08-12-44}          & \textbf{2021-04-29-15-55} & Mean \\ 
                        &                     &                 &  \textbf{[Overcast, Snow]} & \textbf{[Overcast, Snow, Snowing]} & \textbf{[Sun]}            & \textbf{[Overcast, Rain]} &      \\ \hline \hline

            HERO\cite{burnett2021radar}                  &\cite{burnett2021radar}   &   S      &   -           &\e{1.98}/0.53  &   -           &   -           &   -          \\
            FD-RIO (Ours)                                &-                         &   D      &   2.04/0.02   & 2.15/\e{0.02} &   1.70/0.02   &   2.04/0.02   &   1.98/0.02  \\ \hline

            \multicolumn{8}{l}{Standard KITTI evaluation metrics are used; translational error [\%] / rotational error [deg/100m]. (-) indicates results that are not available or not applicable.}\\
            \multicolumn{8}{l}{(S) stands for Sparse method, (D) is for Dense method.}\\
            \end{tabular}%
            }
\end{center}
\end{table*}

\subsection{Ablation Tests}\label{subseq:ablation}
In this section, we perform ablation tests in order to demonstrate the impact of each component of FD-RIO and verify its positive effect on the overall algorithm.  We used the sequence (\textit{2021-04-08-12-44}) from the \textit{Boreas} dataset and the results of these tests are summarized in Table \ref{tab:ablation}.  We begin our tests with a "Radar Only" variant of FD-RIO, test \#1 in Table \ref{tab:ablation}, where we tested our compact dense radar odometry pipeline described in section \ref{subseq:pc_ro} solely without Kalman filter or measurements from the IMU. The results from this test are obviously not acceptable, however, they are important to motivate complementing the slowly sampled measurements of a scanning radar with the much faster sampling rate of an IMU. Next, we tested IMU + Kalman filter where the latter relied only on linear and angular acceleration measurements from the IMU to update its internal states and generate estimations. The results for this test were very poor as one would expect from an IMU only state estimator because of their well-known drift issues. Test \#3 was for Radar + Kalman filter. The results here are still unacceptable and when compared to test \#1, it can be seen that using the Kalman filter alone without fusing new information did not improve the performance. Test \#4 shows the significant improvement achieved when fusing both sensors (i.e., IMU and radar) together using the Kalman filter, it clearly shows the efficacy of this fusion scheme and how sensors of complementing nature (rich and low FPS scanning radar with noisy but fast IMU data) can have poor performance when used individually, but can work in synergy when fused together. Test \#5 shows that the accuracy can be further improved by compensating for the inherent gyro bias in the IMU sensor, a very common practice when dealing with IMUs. Additionally, more improvements can be achieved when passing radar scans through an HPF which is demonstrated in test \#6. This is attributed to the fact that the high-pass filter will sharpen the edges in radar scans which in turn improves the scan registration process. Finally, the best results attained are reported in test \#7 where all components from previous tests are integrated together, this represents the final FD-RIO composition and the results of this test are what is reported in Table \ref{tab:boreas_results}. 

\begin{table}
    \begin{center}
    \caption{The effect of FD-RIO's different components on the translational and rotational errors}\label{tab:ablation}
    \begin{tabular}{ l | c  }\hline
                                            &  \textbf{Boreas}            \\ 
     \textbf{Tests}                          & \textbf{2021-04-08-12-44}   \\\hline \hline
    (\#1) Radar Only                        &       122.15/19.42            \\ 
    (\#2) IMU + KF                          &       588.98/0.04            \\
    (\#3) Radar + KF                        &       122.18/19.40           \\ 
    (\#4) IMU, Radar, and KF                &       6.80/0.05             \\ 
    (\#5) \#4 \& gyro bias                  &       6.77/0.04             \\ 
    (\#6) \#4 \& HPF                        &       1.81/0.04             \\ 
    (\#7) \#4 \& HPF \& gyro bias (FD-RIO) &       1.70/0.02             \\ \hline
    \multicolumn{2}{l}{Standard KITTI evaluation metrics; translational error [\%] /}\\
    \multicolumn{2}{l}{ rotational error [deg/100m].}\\
    \end{tabular}       
    \end{center}
\end{table}

\subsection{Runtime}\label{subseq:runtime}
Although it is challenging to compare the runtime requirement of methods that were developed and evaluated using different environments and hardware setups, we attempt to put runtime capabilities into perspective in Table \ref{tab:hardware_runtime}. This table summarizes the hardware used to test each method from Table \ref{tab:mulran_results} as reported in their respective sources. We recognize that there could be various other factors affecting the timing of algorithms and that the comparison provided in this table might not give a clear cut judgment; however, we believe that Table \ref{tab:hardware_runtime} can serve as a good indicator of which methods can be described as heavy and which are more hardware-friendly.  

\begin{figure}[!t]
    \centering
    \subfloat[]{\includegraphics[width=1.7in]{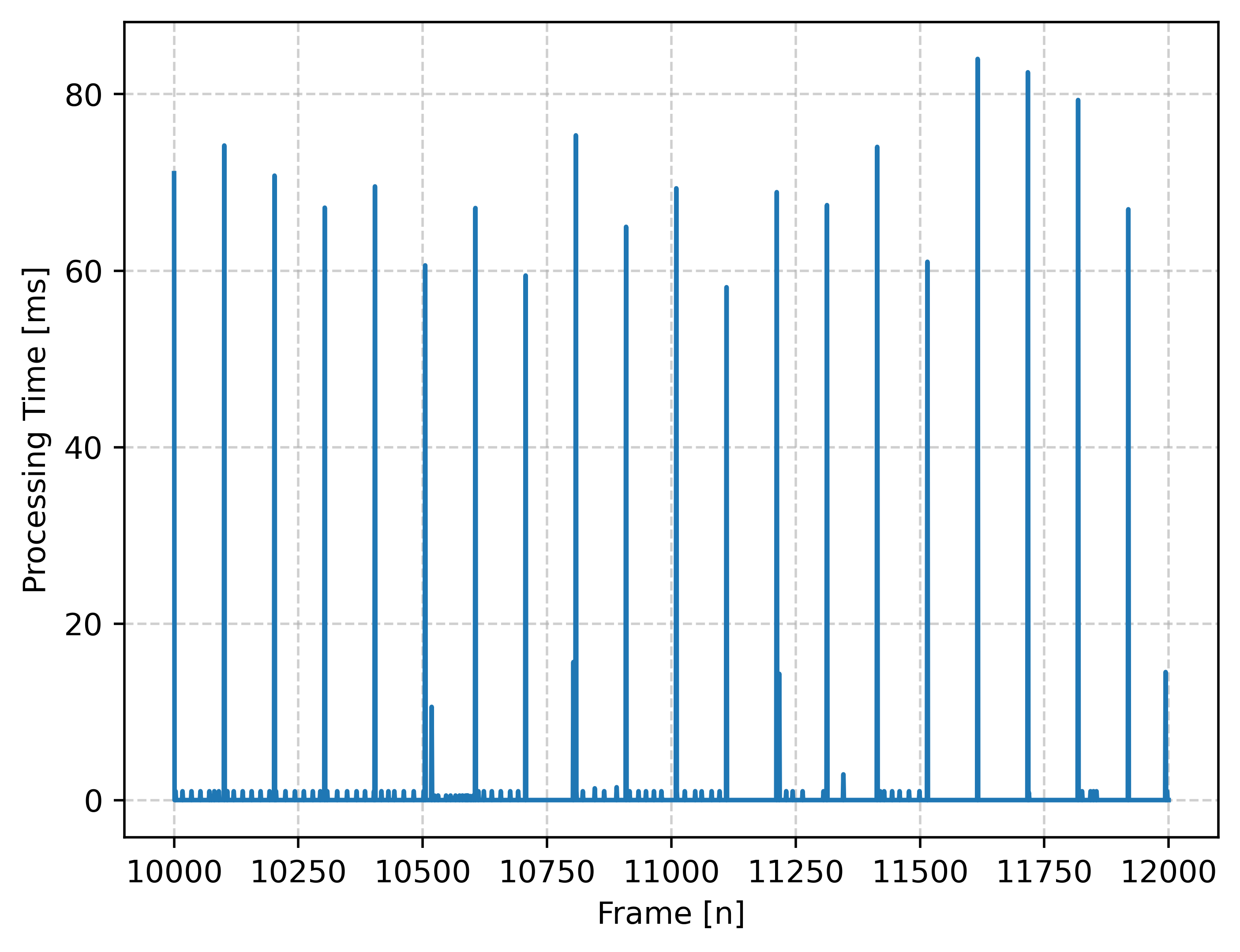}\label{fig:runtime_a}}
    \hfil
    \subfloat[]{\includegraphics[width=1.7in]{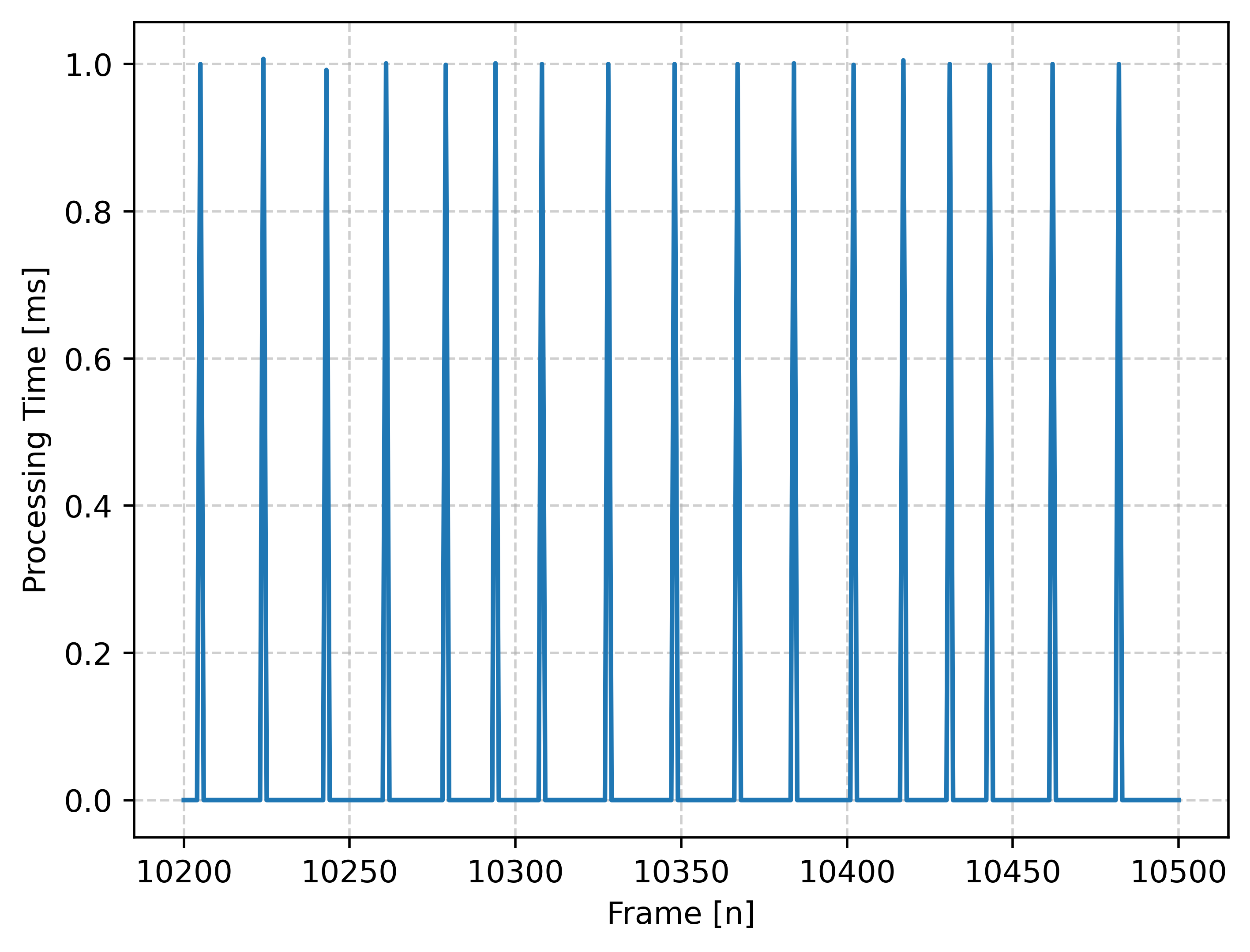}\label{fig:runtime_b}}
    \caption{A snapshot of processing time taken from sequence \textit{2021-04-08-12-44}. Timing includes loading/reading of data and performing \textit{predict} + \textit{update} steps. (a) Frame processing time from both sensors. (b) Frame processing time from IMU only.}\label{fig:runtime}
\end{figure}

FD-RIO was tested using an office PC with Intel(R) i7-13700 2.10 GHz and 32GB of RAM, without utilizing GPUs or multithreading; therefore, it should be practically feasible to run in real-time using a typical embedded system commonly used on mobile platforms. Fig. \ref{fig:runtime_a} shows a snapshot of the time required to process every \textit{update} + \textit{predict} cycle by FD-RIO. The figure shows that there are two groups of spikes. The higher ones are the time needed to process radar scans and they average to 70 ms approximately which translates to 14.3 Hz. The shorter spikes, zoomed in in Fig. \ref{fig:runtime_b}, are the processing time of IMU measurements and they take around 1 ms (i.e., 1000 Hz) to complete. While processing radar scans, a few IMU readings will be lost and their number will depend on the IMU sampling rate; however, those few data points are not causing a noticeable effect on the final estimates.

\begin{figure}[!t]
    \centering
    \subfloat[]{\includegraphics[width=3.5in]{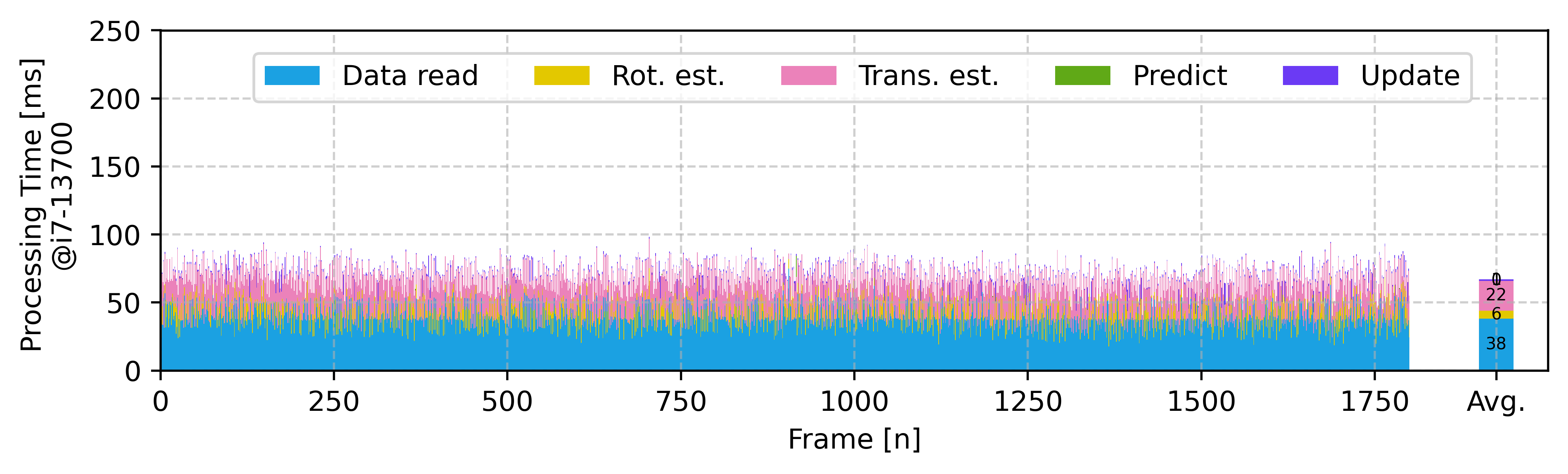}\label{fig:runtime_detailed_a}}
    \vfil
    \subfloat[]{\includegraphics[width=3.5in]{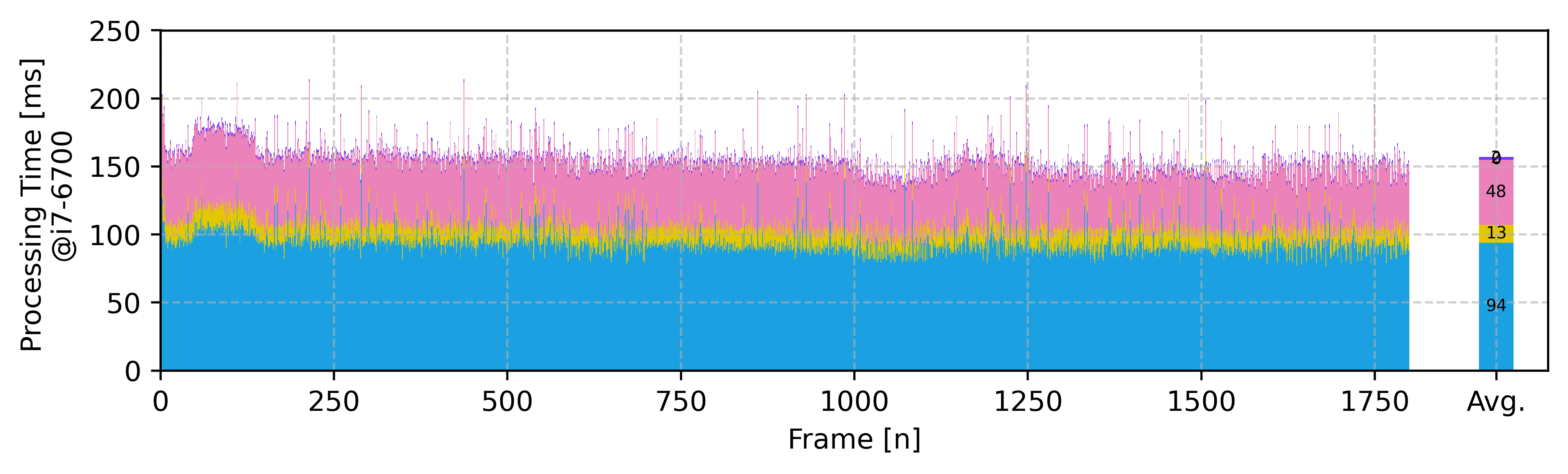}\label{fig:runtime_detailed_b}}
    \caption{A breakdown of the time needed for each of the main steps in FD-RIO while processing radar scans in sequence \textit{2021-04-08-12-44} from the \textit{Boreas} dataset on two different platforms. These steps are loading of data, rotation estimation, translation estimation, \textit{predict}, and \textit{update} steps. (a) Processing time measured on a typical office PC. (b) Processing time measured on an older CPU, an i7-6700.}\label{fig:runtime_detailed}
\end{figure}

Fig. \ref{fig:runtime_detailed_a} shows a breakdown of time needed to process each radar scan through the main steps of FD-RIO. These steps are loading of data, rotation estimation, translation estimation, performing a \textit{predict} step, and  performing an \textit{update} step. The figure shows that the time needed for the \textit{predict} and \textit{update} steps is negligible. The most time consuming step in our setup was the loading of data, followed by the translation and rotation estimation steps which require an average of 22 ms and 6 ms respectively. FD-RIO could be sped up even further by utilizing multiprocessing or multithreading where, for example, data loading and data processing can take place on separate threads. It can also be sped up by utilizing a GPU which can reduce the time needed for the FFT and HPF calculations. Nevertheless, our implementation targets mobile vehicles with limited computational and energy resources and we tried to optimize our algorithm for such scenarios. Finally, in order to marginalize the effect of using a relatively newer CPU compared to methods that were introduced by older publications (e.g., ORORA on i7 6th generation and CFEAR on i7 8th generation); we ran our tests on a significantly older CPU, an i7 6700. For this test, FD-RIO ran at 7.5 Hz for radar scans and 1 kHz for IMU measurements. Considering the results from both Table \ref{tab:hardware_runtime} and Table \ref{tab:mulran_results}, it is evident that even when tested on a much older processor, FD-RIO is at the top of dense radar odometry methods and it is still on par with the state-of-the-art sparse radar odometry methods in terms of both accuracy and runtime capability.

\begin{table*}
    \begin{center}
    \caption{Indicative comparison of the setups used and runtime capabilities of the State-of-the-Art methods}
    \label{tab:hardware_runtime}
    \resizebox{\linewidth}{!}{%
    \begin{tabular}{ l | c | c  l }\hline
     \textbf{Method}    &  \textbf{Type}      & \textbf{FPS}   & \textbf{Hardware} \\\hline \hline

    ORORA \cite{lim2023orora}                      &  S   & 11.5 Hz      & Intel(R) Core i7-7700K                                                    \\
    RadarSLAM (odometry) \cite{hong2022radarslam}  &  S   & 8 Hz         & 2.60 GHz CPU and 16 GB RAM                                                \\
    CFEAR-3 \cite{adolfsson2023cfear}              &  S   & 44 Hz        & Intel i7-8700 k CPU                                                       \\
    CFEAR-3-s50 \cite{adolfsson2023cfear}          &  S   & 5 Hz         & Intel i7-8700 k CPU                                                       \\
    MC-RANSAC \cite{burnett2021mcransac}           &  S   & ~11 Hz       & Quad-core Intel Xeon E3-1505M 3.0 GHz CPU with 32 GB of RAM               \\
    SDRO \cite{zhang2023sdr}                       &  S   & 6$\sim$7Hz   & Intel(R) Core i7-10750H 2.60GHz and 16 GB RAM                             \\ \hline
    $R^3O$ \cite{r3o}                              &  D   & 4.57 Hz      & 2-Intel E5-2695 and 2-Nvidia Quadro RTX-4000, multi-threaded              \\
    FD-RIO (Ours)                                  &  D   & 14.3 Hz (radar), 1 kHz (IMU)             & Intel(R) i7-13700 2.10 GHz and 32GB of RAM \\ \hline
    \multicolumn{4}{l}{(S) stands for Sparse method, (D) is for Dense method.}\\
    \end{tabular}%
    }
    \end{center}
\end{table*}

\section{Conclusion}\label{sec:conclusion}
We have presented FD-RIO, a dense radar inertial odometry algorithm that leverages a well-known states estimator (i.e., Kalman filter) and takes advantage of the complementary nature of different sensors; higher refresh rate of IMU and higher fidelity of scanning radar measurements, it asynchronously updates from both sensors and performs prediction after each update step. Our radar odometry pipeline is based on phase correlation, a technique borrowed from image registration and proved to be successful for estimating pose changes between spinning radar scans. We tested our method using two publicly available datasets and reported performance that is on par with the state-of-the-art, although FD-RIO does not outperform the state-of-the-art on all test sequences, its accuracy was always competitive with the top results, especially in terms of rotation estimation. Taking FD-RIO's unmatched compactness into consideration; our method stands as a very attractive solution for mobile platforms with realistic computational resources. FD-RIO can be further improved by incorporating covariance data from IMU sensors, however, this data was not made available with \textit{MulRan} and \textit{Boreas} datasets. Finally, we plan to formulate uncertainties associated with our odometry pipeline and feed it to the state estimator for a better overall uncertainty estimation.

\IEEEtriggeratref{19}
\bibliographystyle{IEEEtran}
\bibliography{myref}

\begin{thebibliography}{10}
\providecommand{\url}[1]{#1}
\csname url@samestyle\endcsname
\providecommand{\newblock}{\relax}
\providecommand{\bibinfo}[2]{#2}
\providecommand{\BIBentrySTDinterwordspacing}{\spaceskip=0pt\relax}
\providecommand{\BIBentryALTinterwordstretchfactor}{4}
\providecommand{\BIBentryALTinterwordspacing}{\spaceskip=\fontdimen2\font plus
\BIBentryALTinterwordstretchfactor\fontdimen3\font minus
  \fontdimen4\font\relax}
\providecommand{\BIBforeignlanguage}[2]{{%
\expandafter\ifx\csname l@#1\endcsname\relax
\typeout{** WARNING: IEEEtran.bst: No hyphenation pattern has been}%
\typeout{** loaded for the language `#1'. Using the pattern for}%
\typeout{** the default language instead.}%
\else
\language=\csname l@#1\endcsname
\fi
#2}}
\providecommand{\BIBdecl}{\relax}
\BIBdecl

\bibitem{NaderSurvey}
N.~J. Abu-Alrub and N.~A. Rawashdeh, ``Radar odometry for autonomous ground
  vehicles: A survey of methods and datasets,'' \emph{IEEE Transactions on
  Intelligent Vehicles}, vol.~9, no.~3, pp. 4275--4291, 2024.

\bibitem{cen2018precise}
S.~H. Cen and P.~Newman, ``Precise ego-motion estimation with millimeter-wave
  radar under diverse and challenging conditions,'' in \emph{2018 IEEE
  International Conference on Robotics and Automation (ICRA)}.\hskip 1em plus
  0.5em minus 0.4em\relax IEEE, 2018, pp. 6045--6052.

\bibitem{cen2019radar}
S.~Cen and P.~Newman, ``Radar-only ego-motion estimation in difficult settings
  via graph matching,'' in \emph{2019 International Conference on Robotics and
  Automation (ICRA)}.\hskip 1em plus 0.5em minus 0.4em\relax IEEE, 2019, pp.
  298--304.

\bibitem{burnett2021mcransac}
K.~Burnett, A.~P. Schoellig, and T.~D. Barfoot, ``Do we need to compensate for
  motion distortion and doppler effects in spinning radar navigation?''
  \emph{IEEE Robotics and Automation Letters}, vol.~6, no.~2, pp. 771--778,
  2021.

\bibitem{burnett2021radar}
K.~Burnett, D.~J. Yoon, A.~P. Schoellig, and T.~D. Barfoot, ``Radar odometry
  combining probabilistic estimation and unsupervised feature learning,''
  \emph{arXiv preprint arXiv:2105.14152}, 2021.

\bibitem{hong2022radarslam}
Z.~Hong, Y.~Petillot, A.~Wallace, and S.~Wang, ``Radarslam: A robust
  simultaneous localization and mapping system for all weather conditions,''
  \emph{The international journal of robotics research}, vol.~41, no.~5, pp.
  519--542, 2022.

\bibitem{adolfsson2023cfear}
D.~Adolfsson, M.~Magnusson, A.~Alhashimi, A.~J. Lilienthal, and H.~Andreasson,
  ``Lidar-level localization with radar? the cfear approach to accurate, fast,
  and robust large-scale radar odometry in diverse environments,'' \emph{IEEE
  Transactions on Robotics}, vol.~39, no.~2, pp. 1476--1495, 2023.

\bibitem{zhang2023sdr}
R.~Zhang, Y.~Zhang, D.~Fu, and K.~Liu, ``Scan denoising and normal distribution
  transform for accurate radar odometry and positioning,'' \emph{IEEE Robotics
  and Automation Letters}, vol.~8, no.~3, pp. 1199--1206, 2023.

\bibitem{lim2023orora}
H.~Lim, K.~Han, G.~Shin, G.~Kim, S.~Hong, and H.~Myung, ``Orora: Outlier-robust
  radar odometry,'' in \emph{2023 IEEE International Conference on Robotics and
  Automation (ICRA)}.\hskip 1em plus 0.5em minus 0.4em\relax IEEE, 2023, pp.
  2046--2053.

\bibitem{lu2020milliego}
C.~X. Lu, M.~R.~U. Saputra, P.~Zhao, Y.~Almalioglu, P.~P. De~Gusmao, C.~Chen,
  K.~Sun, N.~Trigoni, and A.~Markham, ``milliego: single-chip mmwave radar
  aided egomotion estimation via deep sensor fusion,'' in \emph{Proceedings of
  the 18th Conference on Embedded Networked Sensor Systems}, 2020, pp.
  109--122.

\bibitem{barnes2019masking}
D.~Barnes, R.~Weston, and I.~Posner, ``Masking by moving: Learning
  distraction-free radar odometry from pose information,'' \emph{arXiv preprint
  arXiv:1909.03752}, 2019.

\bibitem{weston2022fastMbyM}
R.~Weston, M.~Gadd, D.~De~Martini, P.~Newman, and I.~Posner, ``Fast-mbym:
  Leveraging translational invariance of the fourier transform for efficient
  and accurate radar odometry,'' in \emph{2022 International Conference on
  Robotics and Automation (ICRA)}, 2022, pp. 2186--2192.

\bibitem{checchin2010radar}
P.~Checchin, F.~G{\'e}rossier, C.~Blanc, R.~Chapuis, and L.~Trassoudaine,
  ``Radar scan matching slam using the fourier-mellin transform,'' in
  \emph{Field and Service Robotics: Results of the 7th International
  Conference}.\hskip 1em plus 0.5em minus 0.4em\relax Springer, 2010, pp.
  151--161.

\bibitem{park2020pharao}
Y.~S. Park, Y.-S. Shin, and A.~Kim, ``Pharao: Direct radar odometry using phase
  correlation,'' in \emph{2020 IEEE International Conference on Robotics and
  Automation (ICRA)}.\hskip 1em plus 0.5em minus 0.4em\relax IEEE, 2020, pp.
  2617--2623.

\bibitem{r3o}
D.~L.~S. Lubanco, A.~Hashem, M.~Pichler-Scheder, A.~Stelzer, R.~Feger, and
  T.~Schlechter, ``R3o: Robust radon radar odometry,'' \emph{IEEE Transactions
  on Intelligent Vehicles}, vol.~9, no.~1, pp. 231--246, 2024.

\bibitem{fft_reddy}
B.~Reddy and B.~Chatterji, ``An fft-based technique for translation, rotation,
  and scale-invariant image registration,'' \emph{IEEE Transactions on Image
  Processing}, vol.~5, no.~8, pp. 1266--1271, 1996.

\bibitem{almalioglu2020milli}
Y.~Almalioglu, M.~Turan, C.~X. Lu, N.~Trigoni, and A.~Markham, ``Milli-rio:
  Ego-motion estimation with low-cost millimetre-wave radar,'' \emph{IEEE
  Sensors Journal}, vol.~21, no.~3, pp. 3314--3323, 2020.

\bibitem{de2023novel}
P.~R.~M. de~Araujo, M.~Elhabiby, S.~Givigi, and A.~Noureldin, ``A novel method
  for land vehicle positioning: Invariant kalman filters and
  deep-learning-based radar speed estimation,'' \emph{IEEE Transactions on
  Intelligent Vehicles}, vol.~8, no.~9, pp. 4275--4286, 2023.

\bibitem{holder2019real}
M.~Holder, S.~Hellwig, and H.~Winner, ``Real-time pose graph slam based on
  radar,'' in \emph{2019 IEEE Intelligent Vehicles Symposium (IV)}.\hskip 1em
  plus 0.5em minus 0.4em\relax IEEE, 2019, pp. 1145--1151.

\bibitem{liang2020scalable}
Y.~Liang, S.~M{\"u}ller, D.~Schwendner, D.~Rolle, D.~Ganesch, and I.~Schaffer,
  ``A scalable framework for robust vehicle state estimation with a fusion of a
  low-cost imu, the gnss, radar, a camera and lidar,'' in \emph{2020 IEEE/RSJ
  International Conference on Intelligent Robots and Systems (IROS)}.\hskip 1em
  plus 0.5em minus 0.4em\relax IEEE, 2020, pp. 1661--1668.

\bibitem{welch1995introduction}
G.~Welch, ``An introduction to the kalman filter,'' 1995.

\bibitem{chen2011kalman}
S.-Y. Chen, ``Kalman filter for robot vision: a survey,'' \emph{IEEE
  Transactions on industrial electronics}, vol.~59, no.~11, pp. 4409--4420,
  2011.

\bibitem{mulran_dataset}
G.~Kim, Y.~S. Park, Y.~Cho, J.~Jeong, and A.~Kim, ``Mulran: Multimodal range
  dataset for urban place recognition,'' in \emph{2020 IEEE International
  Conference on Robotics and Automation (ICRA)}, 2020, pp. 6246--6253.

\bibitem{boreas_dataset}
K.~Burnett, D.~J. Yoon, Y.~Wu, A.~Z. Li, H.~Zhang, S.~Lu, J.~Qian, W.-K. Tseng,
  A.~Lambert, K.~Y. Leung, A.~P. Schoellig, and T.~D. Barfoot, ``Boreas: A
  multi-season autonomous driving dataset,'' \emph{The International Journal of
  Robotics Research}, vol.~42, no. 1-2, pp. 33--42, 2023.

\bibitem{KITTI}
A.~Geiger, P.~Lenz, and R.~Urtasun, ``Are we ready for autonomous driving? the
  kitti vision benchmark suite,'' in \emph{2012 IEEE Conference on Computer
  Vision and Pattern Recognition}, 2012, pp. 3354--3361.

\bibitem{what_should_be_learnt}
H.~Zhan, C.~S. Weerasekera, J.-W. Bian, and I.~Reid, ``Visual odometry
  revisited: What should be learnt?'' in \emph{2020 IEEE International
  Conference on Robotics and Automation (ICRA)}, 2020, pp. 4203--4210.

\bibitem{oxford_radar_dataset}
D.~Barnes, M.~Gadd, P.~Murcutt, P.~Newman, and I.~Posner, ``The oxford radar
  robotcar dataset: A radar extension to the oxford robotcar dataset,'' in
  \emph{2020 IEEE International Conference on Robotics and Automation (ICRA)},
  2020, pp. 6433--6438.

\end{thebibliography}

\end{document}